%% file: xMain.tex
\RequirePackage{fix-cm}
\documentclass{article} %
\usepackage{iclr2027_conference,times}

\input{wMacros}
\input{math_commands}

\usepackage{lipsum}
\usepackage{hyperref}
\usepackage{url}
\usepackage{enumitem}
\usetikzlibrary{calc}
\usetikzlibrary{backgrounds}

\usepackage{amsmath}
\usepackage{amsthm}
\usepackage{amssymb}
\usepackage{mathtools}
\usepackage{amsfonts}
\usepackage{sansmath}

\usepackage{booktabs}
\usepackage{siunitx}

\usepackage[table]{xcolor}
\usepackage{subcaption}
\usepackage{wrapfig}
\usepackage{tikz}
\usetikzlibrary{arrows.meta}
\usepackage{graphicx}
\usepackage{pgfplots}
\pgfplotsset{compat=1.18}

\usepackage[most]{tcolorbox}

\definecolor{contribblue}{HTML}{0072B2}
\definecolor{contribgold}{HTML}{FCA636}

\newtcolorbox{contribbox}[1][]{
  enhanced,
  title=#1,
  colback=contribgold!10,
  colframe=contribblue,
  colbacktitle=contribgold!5,
  coltitle=contribblue,
  left=0pt, right=8pt, top=10pt, bottom=2pt,
  attach boxed title to top left={xshift=15pt,yshift=-10pt},
  boxed title style={
    frame hidden,
    colback=contribgold!45
  },
  rounded corners,
  arc=3pt
}

\title{Outperforming 1000 Euclidean Layers With Hyperbolic Goal-Conditioned RL}

\author{%
Timo Klein$^{\star,1,2}$, Thomas Lang$^{\star,1,2}$, Andrii Shkabrii$^{1,2}$, Lukas Miklautz$^{3}$, Claudia Plant$^{1,4}$ \\
\bf Yllka Velaj$^{1,4}$, Sebastian Tschiatschek$^{1,4}$ \\
\normalfont $^{1}$ Faculty of Computer Science, University of Vienna, Vienna, Austria \\
\normalfont $^{2}$ UniVie Doctoral School Computer Science, University of Vienna, Vienna, Austria \\
\normalfont $^{3}$ Department of Machine Learning and Systems Biology, Max Planck Institute of Biochemistry, \\
\normalfont Martinsried, Germany \\
\normalfont $^{4}$ ds:UniVie, University of Vienna, Vienna, Austria \\
\normalfont $^{\star}$ Joint first authors, 
\normalfont \texttt{firstname.lastname@univie.ac.at}
}

\makeatletter
\newlength{\abstractgap}
\renewenvironment{abstract}{\vskip.075in\centerline{\large\sc Abstract}%
  \list{}{\rightmargin\leftmargin \topsep=\abstractgap \partopsep=\z@ \parskip=\z@}\item\relax}
  {\par\endlist\vskip 1ex}
\makeatother

\iclrfinalcopy %
\begin{document}

\vspace*{-11mm} %
\maketitle
\vspace*{-2mm} %
\lhead{Preprint.}%
\vspace{-5mm}
\begin{abstract}
Scaling self-supervised contrastive reinforcement learning (CRL) to 1000-layer Euclidean neural networks has enabled state-of-the-art goal-reaching performance, yet wall-clock time scales linearly with depth. 
We find that changing the geometry achieves stronger performance with far shallower networks.
This matters in complex navigation environments, where walls can make route distance deviate substantially from Euclidean distance. In a U-shaped maze, we show that the longer the route, the more it deviates from Euclidean distance.
Hyperbolic CRL addresses this mismatch by replacing Euclidean distances in CRL's contrastive objective with geodesic distances in hyperbolic space.
On hard navigation tasks, inductive bias beats scale: depth 64 hyperbolic agents outperform 1024-layer-deep Euclidean networks while training more than $12 \times$ faster and using about $13 \times$ fewer parameters. We observe that hyperbolic agents behave qualitatively better in hard humanoid mazes, where they follow corridors to reach goals, whereas Euclidean agents try straight-line shortcuts over walls.
Our results show that matching the representation geometry to the task can substitute for substantial network depth. Code is at \href{https://github.com/Probabilistic-and-Interactive-ML/HypGCRL}{https://github.com/Probabilistic-and-Interactive-ML/HypGCRL}.
\end{abstract}

\section{Introduction}\label{sec:introduction}

\begin{wrapfigure}{r}{0.5\linewidth}
        \centering
        \vspace{-20mm}
        \includegraphics[width=\linewidth]{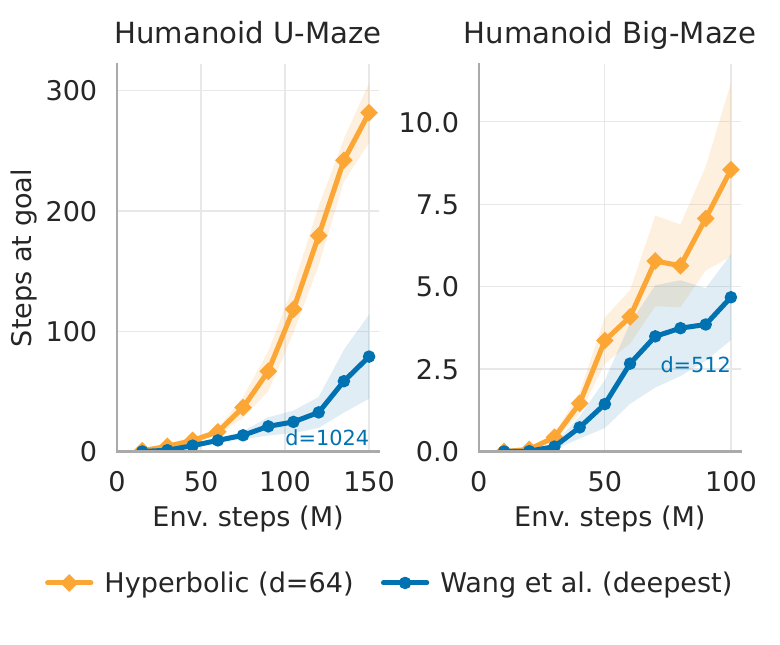}
        \vspace{-10mm}
        \caption{\textbf{Hyperbolic agents outperform much deeper Euclidean ones.} A hyperbolic agent of depth 64 reaches a higher steps at goal than the depth 1024 Euclidean agent while using $12 \times$ less wall-clock time on Humanoid U-Maze. Euclidean results are taken from \citet[Figure 12]{Wang2025ScalingCRL}, we use their step budget. Wall-clock times are reported in Appendix~\ref{app:wall_clock}.}
        \label{fig:first_page}
        \vspace{-5mm}
\end{wrapfigure}

Consider a humanoid having to find the goal in a maze. Navigating purely based on straight-line distance would lead this agent down many dead ends, because straight-line distance is a misleading measure of how far a goal is in the maze. In this setting, the \textcolor{contribblue}{\emph{straight-line distance}} between two positions ignores walls, whereas the \textcolor{contribgold}{\emph{route distance}} is the length of the actual path between a state and a goal. Goal-conditioned RL (GCRL) trains agents to reach arbitrary goals without a hand-designed reward. Contrastive reinforcement learning (CRL)~\citep{Eysenbach2022ContrastiveRL} casts GCRL as a classification problem and uses an InfoNCE objective~\citep{Oord2018InfoNCE} to predict whether a goal is reachable from a state-action pair, based on the Euclidean distance between their embeddings~\citep{Bortkiewicz2025AcceleratingCRL}. Euclidean embeddings make straight-line proximity easy to represent, and this proxy works at short range but misleads once walls are present. \citet[Figure~9]{Wang2025ScalingCRL} show that on Ant U4-Maze, Euclidean agents learn to follow route structure only when using deep networks.

This reliance on depth is expensive. Recent work scales CRL with GPU-accelerated simulation~\citep{Bortkiewicz2025AcceleratingCRL} and networks up to 1024 layers deep~\citep{Wang2025ScalingCRL}. However, these Euclidean agents perform best at extreme depth, making training very expensive (Figure~\ref{fig:first_page}). \citet[Section~4.4]{Wang2025ScalingCRL} evaluate 1024-layer networks on a single task, Humanoid U-Maze, and report that computational constraints kept them from scaling further. Our Ant U5-Maze study shows why: Euclidean representations must learn to \quoteeng{unroll} the maze to reflect route distance, which can require substantial depth. We therefore hypothesize that deep Euclidean agents use part of their capacity to compensate for the mismatch between the geometry of their embeddings and the geometry of the environment. This raises our paper's central question: \emph{Is Euclidean distance the right geometry for goal-reaching?}

Hyperbolic space offers a natural alternative. It is approximately Euclidean at small scales but tree-like at large scales, and trees embed in it with arbitrarily low distortion~\citep{Sarkar2011LowDistortion}. Existing hyperbolic RL methods~\citep{Cetin2023HyperbolicDeepRL, Klein2026HyperPP}, however, assign no meaning to embedding distance and learn hierarchy only indirectly through their standard RL objectives. Quasimetric approaches~\citep{Wang2022IntervalQuasimetric, Liu2023MetricResidualNetwork, Wang2023OptimalGCRLQuasimetric, Myers2024LearningTemporalDistances} use asymmetric distances but retain their Euclidean embedding geometry. Consequently, quasimetric CRL agents perform poorly on mazes~\citep[Appendix~A.4]{Wang2025ScalingCRL}. To our knowledge, no prior work has adopted hyperbolic spaces for improving CRL.

Our work investigates this change. Hyperbolic CRL keeps the contrastive objective but replaces Euclidean distances of embedded points with geodesic distances, at a small, fixed computational cost. We motivate this choice with a case study in Ant U5-Maze. Locally, route distances in this environment coincide with straight-line distances, while globally they become increasingly tree-like. Hyperbolic space matches this metric structure. At equal depth, hyperbolic agents outperform Euclidean agents on most tasks. On the Humanoid mazes, they beat Euclidean agents with many times deeper networks. These results suggest that a better choice of representation geometry can substitute for substantial network depth.

\begin{contribbox}[\textbf{Key Contributions}]
    \begin{enumerate}[leftmargin=0.7cm]
        \item \textbf{\textcolor{contribblue}{Hyperbolic CRL.}} We introduce a drop-in modification to CRL that projects embeddings onto hyperbolic space and estimates goal reachability by geodesic distance. This gives the agent an inductive bias toward tree-like route structure at a small computational cost.

        \item \textbf{\textcolor{contribblue}{Geometry of goal-reaching.}} Section~\ref{sec:analysis} shows that maze route distances are approximately Euclidean at short range and increasingly tree-like at long range. Hyperbolic space captures this structure more faithfully than Euclidean geometry, losing its straight-line bias.

        \item \textbf{\textcolor{contribblue}{A better performance-efficiency frontier.}} At fixed depth, hyperbolic agents outperform Euclidean agents on most tasks and stitch long routes more reliably. Figure~\ref{fig:first_page} shows depth 64 hyperbolic agents outperforming Euclidean baselines up to $16\times$ deeper while training in at least $7\times$ less wall-clock time.
    \end{enumerate}
\end{contribbox}

\section{Background}\label{sec:background}

We model goal reaching~\citep{Kaelbling1993AchieveGoals, Andrychowicz2017HER} as a goal-conditioned MDP $\mathcal{M}_g = (\statespace, \actionspace, p_0, p, p_g, r_g, \gamma)$. The agent starts in a state $s_0 \sim p_0$. At each step $t$, it observes a state $s_t \in \statespace$, takes an action $a_t \in \actionspace$, and moves to $s_{t+1} \sim p(\cdot \mid s_t, a_t)$. Goals $g \in \mathcal{G}$ are sampled from $p_g$, and a function $f: \mathcal{S} \to \mathcal{G}$ extracts the goal-relevant part of a state, such as the agent's $(x, y)$ position~\citep{Wang2025ScalingCRL}. We write $g_t \coloneqq f(s_t)$ for the goal achieved at step $t$. Following \citet{Eysenbach2022ContrastiveRL}, the reward is the density of achieving $g$ at the next step, $r_g(s_t, a_t) \coloneqq (1 - \gamma)\, p(g_{t+1} = g \mid s_t, a_t)$, with discount factor $\gamma \in [0,1)$. A goal-conditioned policy $\pi_g(a \mid s) \coloneqq \pi(a \mid s, g)$ with commanded goal $g$ is trained to maximize the expected return $\mathbb{E}_{p_0, p_g, \pi_g}\big[\sum_{t=0}^{\infty} \gamma^{t} r_g(s_t, a_t)\big]$. Let $p^{\pi_g}$ denote densities under trajectories generated by the policy $\pi_g$. Then, the $Q$-function of $\pi_g$ is a density over goals:\begin{equation}
    Q^{\pi}_g(s, a) = p^{\pi_g}_{\gamma}(g \mid s, a) \coloneqq (1 - \gamma) \sum_{k=0}^{\infty} \gamma^{k}\, p^{\pi_g}(g_{t+k+1} = g \mid s_t = s, a_t = a).
\end{equation}
Thus, $Q^{\pi}_g(s, a)$ is the discounted visitation density of the policy at the goal.\\
\textbf{Contrastive Reinforcement Learning.}
Contrastive Reinforcement Learning (CRL) estimates this density, averaged over goals, via classification rather than temporal-difference (TD) learning~\citep{Eysenbach2022ContrastiveRL}. CRL uses two encoders $\phi \colon \mathcal{S} \times \mathcal{A} \to \mathcal{Z}$ and $\psi \colon \mathcal{G} \to \mathcal{Z}$ that embed into a shared representation space $\mathcal{Z}$. An energy is computed using the similarity between the two embeddings:
\begin{equation}\label{eq:energy}
  e (s, a, g) \;\coloneqq\; \, \operatorname{sim} \bigl(\phi (s, a),\;\psi (g)\bigr)
\end{equation}
Here $\operatorname{sim} \colon \mathcal{Z} \times \mathcal{Z} \to \mathbb{R}$ is a similarity function on $\mathcal{Z}$, e.g., a negative distance~\citep{Bortkiewicz2025AcceleratingCRL}, the dot product, or the cosine similarity. CRL is an actor-critic method: its critic is the energy $e$ and its actor is the policy $\pi_\theta(a \mid s, g)$ with parameters $\theta$. The policy is then trained to maximize $e$. The critic is trained with the InfoNCE objective~\citep{Oord2018InfoNCE, Bortkiewicz2025AcceleratingCRL} over minibatches of $(s_t, a_t, g_{t+k})$ tuples, where the positive sample $g_{t+k}$ is the goal achieved at a future step $t + k$ of the same trajectory, with $k \sim \mathrm{Geom}(1 - \gamma)$ on $\{1, 2, \dots\}$. Achieved goals from other trajectories in the minibatch are treated as negative samples. The implementation we build on uses the negative Euclidean distance~\citep{Bortkiewicz2025AcceleratingCRL} as $\operatorname{sim}$. This paper replaces it with the negative geodesic distance in hyperbolic space.\\
\textbf{Hyperbolic geometry.}
Hyperbolic space is a simply connected Riemannian manifold of constant negative sectional curvature $-c$, with curvature magnitude $c > 0$. In this work we use the \emph{hyperboloid} (Lorentz) model, which represents hyperbolic space as the forward sheet of a two-sheeted hyperboloid $\hyperboloid = \{\bm{x} \in \mathbb{R}^{n+1} : \minner{\bm{x}}{\bm{x}} = -1/c,\; x_0 > 0\}$ in Minkowski space~\citep{lee2018introduction}. Here, $\minner{\bm{x}}{\bm{y}} \coloneqq -x_0 y_0 + \sum_{i=1}^{n} x_i y_i$ denotes the Minkowski inner product.\\
\textbf{Hyperbolic encoding.}
The output of a Euclidean encoder $\bm{u} \in \mathbb{R}^n$ is mapped to the hyperboloid in two steps~\citep{Nickel2018LearningContinuousHierarchies}. First, $\bm{u}$ is identified with the vector $\bm{v} = (0, \bm{u})$ of the tangent space $\htangent{\horigin}$ at the origin $\horigin = (1/\sqrt{c}, \bm{0})$. Second, the exponential map $\expmap{\horigin}$ projects $\bm{v}$ onto the manifold. The closed-form expressions are given in Appendix~\ref{app:hyperbolic:hyp_ops} alongside other hyperbolic space operations.

\section{Related Work}\label{sec:related_work}

\textbf{Goal-conditioned reinforcement learning.} Goal-conditioned RL (GCRL) trains a policy to reach any target state by conditioning it on that goal~\citep{Kaelbling1993AchieveGoals, Schaul2015UniversalValue}. Hindsight experience replay~\citep{Andrychowicz2017HER} implemented GCRL by relabeling failed trajectories with the goals they actually reached. We build on Contrastive RL (CRL)~\citep{Eysenbach2022ContrastiveRL}, which trains the critic using an InfoNCE objective~\citep{Oord2018InfoNCE, Bortkiewicz2025AcceleratingCRL} rather than explicitly learning rewards or distances. Our implementations follow \citet{Bortkiewicz2025AcceleratingCRL} for online CRL and OGBench~\citep{Park2025OGBench} for offline CRL, which include recent improvements such as stronger regularization~\citep{Zheng2024StabilizingCRL}. Beyond GCRL, scaling network size has long been difficult in deep RL. Recent work has made them trainable using regularized residual critics~\citep{Nauman2024BRO, Lee2025SimBa, Lee2025SimBaV2} or mixture-of-experts layers~\citep{ObandoCeron2024MoEforRL}. \citet{Wang2025ScalingCRL} adapted these architectures to CRL and trained agents with up to 1024 layers. Our work is complementary: we keep the CRL objective and change the geometry of its representations.\\
\textbf{Quasimetric RL.}
\citet{Bortkiewicz2025AcceleratingCRL} observe that CRL works best when energies satisfy the triangle inequality, e.g., the Euclidean distance outperforms the squared Euclidean distance. This fits the view that temporal distances satisfy the triangle inequality and that goal-reaching distances should enforce it by construction~\citep{Wang2023OptimalGCRLQuasimetric, Myers2024LearningTemporalDistances}. Temporal distances are asymmetric, so prior work builds quasimetrics into both architectures~\citep{Wang2022IntervalQuasimetric, Liu2023MetricResidualNetwork} and goal-reaching algorithms~\citep{Wang2023OptimalGCRLQuasimetric, Myers2024LearningTemporalDistances, Myers2025OfflineQuasimetric, Zheng2026MultiStepQuasimetric}. While hyperbolic geodesic distances also satisfy the triangle inequality, they depart from Euclidean distance in a different way. Quasimetrics are asymmetric but preserve the flat coordinates of Euclidean geometry, whereas hyperbolic distances preserve symmetry while changing the curvature. \citet[Figure 15]{Wang2025ScalingCRL} report that the quasimetric CRL agent of \citet{Myers2024LearningTemporalDistances} performs poorly maze-like tasks, which motivate our work.\\
\textbf{Hyperbolic representation learning.}
Hyperbolic embeddings outperform Euclidean ones on hierarchical data, especially in low dimensions~\citep{Nickel2017PoincareEmbeddings, Nickel2018LearningContinuousHierarchies}, because hyperbolic space embeds trees with arbitrarily low distortion~\citep{Sarkar2011LowDistortion, Sala2018RepresentationTradeOffs}. Since \citet{Ganea2018HyperbolicNNs} introduced hyperbolic neural networks, later work has added linear layers~\citep{Shimizu2021HyperbolicNNpp, Yang2024Hypformer, vanDerKlis2026FGG-LNN} that avoid Riemannian optimization~\citep{Bonnabel2013RiemannianSGD, Becigneul2019RiemannianOptimization}, convolutional layers~\citep{Bdeir2024FullyHyperbolicCNNs}, and self-attention~\citep{Yang2024Hypformer}. Contrastive learning optimizes relative distances, and is thus well-suited for hyperbolic geometry~\citep{Suris2021HypContrastive1, Ge2023HypContrastive2, Mettes2024SurveyHyperbolicCV}, particularly in vision-language alignment~\citep{Desai2023Meru, Mettes2025HycoCLIP}. In deep RL, \citet{Cetin2023HyperbolicDeepRL} introduced hyperbolic representations and argued that MDP states evolve like a tree by measuring the relative $\delta$-hyperbolicity~\citep{Gromov1987Hyperbolicity, Khrulkov2020DeltaRelHyperbolicity} of learned representations. \citet{Klein2026HyperPP} identified why hybrid hyperbolic-Euclidean agents fail to train and stabilized them with regularization, improving over prior hyperbolic agents on ProcGen and Atari. Neither work tests whether the environment itself is tree-like. To our knowledge, we are the first to use hyperbolic representations in goal-conditioned RL.\\

\section{Case Study: Analyzing the Geometry of a Deceptive Maze}\label{sec:analysis}

\begin{figure}[t]
\centering
\captionsetup[subfigure]{font=footnotesize,position=top,skip=1pt,justification=centering,singlelinecheck=false}
\noindent%
\begin{subfigure}[t]{0.32\linewidth}
  \centering
  \caption{$\delta$-hyperbolicity}\label{fig:analysis-fig1-delta}
  \includegraphics[width=\linewidth]{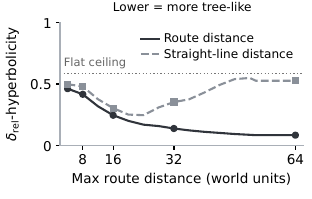}
\end{subfigure}\hfill%
\begin{subfigure}[t]{0.32\linewidth}
  \centering
  \caption{Embedding dist. vs route dist.}\label{fig:analysis-fig1-correlation}
  \includegraphics[width=\linewidth]{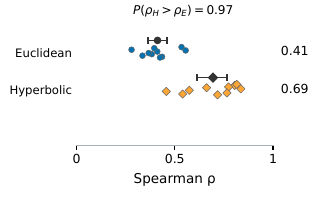}
\end{subfigure}\hfill%
\begin{subfigure}[t]{0.32\linewidth}
  \centering
  \caption{Route ordering (50\% is chance)}\label{fig:analysis-fig1-ordering}
  \includegraphics[width=\linewidth]{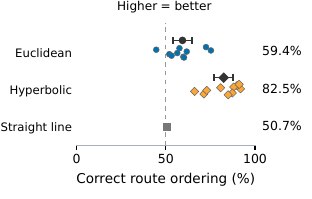}
\end{subfigure}

\caption{\textbf{Hyperbolic critics better reflect the geometric structure of Ant U5-Maze.}
    \textbf{\ref{fig:analysis-fig1-delta})}
    \emph{The maze becomes increasingly more tree-like at longer task horizons under route distance, when walls are taken into account}. We report relative $\delta$-hyperbolicity of Ant U5-Maze (lower is more tree-like) under increasing task horizons.
    \textbf{\ref{fig:analysis-fig1-correlation})}
    \textit{Hyperbolic embedding distances track route distances more closely than Euclidean embedding distances}. $P(\rho_H>\rho_E)$ estimates the probability that a hyperbolic run has a higher Spearman correlation than a Euclidean run.
    \textbf{\ref{fig:analysis-fig1-ordering})} We sample maze locations with similar straight-line distance but different route distance. The percentage shows how often the critic assigns the longer route a larger embedding distance. \emph{The hyperbolic critic guides the agent along the correct route more frequently.} 
    In \ref{fig:analysis-fig1-correlation} and \ref{fig:analysis-fig1-ordering}, we show the 10 seeds per method with means and 95\% bootstrap confidence intervals.}
    \vspace{-5mm}
\label{fig:analysis-geometry-correlations}
\end{figure}

\begin{wrapfigure}{r}{0.60\textwidth}
  \centering
  \vspace{-5mm}
  \captionsetup[subfigure]{
    font=footnotesize,
    position=top,
    skip=1pt,
    justification=centering,
    singlelinecheck=false
  }
  \begin{subfigure}[t]{0.581818\linewidth}
    \centering
    \caption{Value maps}
    \label{fig:analysis-fig2-value-func}
    \includegraphics[width=\linewidth]{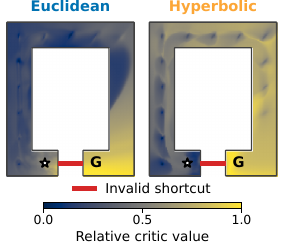}
  \end{subfigure}\hfill%
  \begin{subfigure}[t]{0.412121\linewidth}
    \centering
    \caption{Gradient alignment}
    \label{fig:analysis-fig2-gradient-alignment}
    \includegraphics[width=\linewidth]{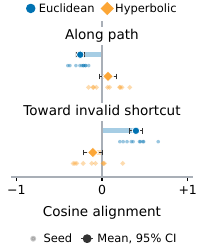}
  \end{subfigure}

  \caption{\textbf{Critic values and gradients in Ant U5-Maze.}
    Both critics have depth 16.
    \textbf{\ref{fig:analysis-fig2-value-func})} \emph{The Euclidean critic assigns high Q-values across the wall near the start, whereas the hyperbolic critic values increase along the route towards the goal (brighter colors)}. $\bm \star$ = start, \textbf{G} = goal, and \textbf{\textcolor{red}{---}} = invalid shortcut.
    \textbf{\ref{fig:analysis-fig2-gradient-alignment})} For state-goal pairs with long routes but short straight-line distances, we measure the cosine similarity between $\nabla_{x,y}Q$, and two directions: along the optimal maze route and straight towards the goal. \emph{Euclidean gradients tend to point towards the wall, while hyperbolic critics do not exhibit this bias}. Results are averaged over 10 seeds.}
  \label{fig:analysis-values-gradients}
\end{wrapfigure}

We study the geometry of goal-reaching in Ant U5-Maze, a complex navigation task in which the agent starts on the left corner of a U-shaped corridor (Figure~\ref{fig:analysis-values-gradients}). In this task, distant goals lie in the opposing right corner, and reaching them requires traversing the entire environment. We measure distances in \textbf{world units}, where one world unit is one unit of the simulator's $(x, y)$ coordinates, and use the following two distances:
\begin{description}[leftmargin=1.2em, itemsep=1pt, topsep=3pt]
  \item[\color{contribgold} Route distance:] length of the shortest path between two positions within the maze.
  \item[\color{contribblue} Straight-line distance:] Euclidean distance between two positions, ignoring walls.
\end{description}
Following Section~\ref{sec:background}, Euclidean and hyperbolic critics use the negative embedding distance as the energy: $Q(s,a,g) = e(s,a,g) = -d\big(\phi(s,a), \psi(g)\big)$, where $d$ is the Euclidean distance or the geodesic distance on the hyperboloid, respectively. We analyze these critics from three perspectives: the geometry of the environment, the learned embedding distances, and the spatial variation in the critic's value. Appendix~\ref{app:u5-case} contains details on how all of the following experiments are implemented.

\paragraph{Is the environment tree-like?}
Intuitively, navigation in a tree-like environment depends on shared routes that branch towards distinct goals. This structure is often difficult to discern locally but becomes clear at global scales. Thus, we examine whether the maze's geometry becomes increasingly tree-like at longer task horizons. Since route distances reflect how different parts of the environment connect, they let us quantify tree-like geometry using Gromov's $\delta$-hyperbolicity~\citep{Gromov1987Hyperbolicity}. This measure quantifies how closely the distances among four given locations in the environment resemble a tree, where a smaller $\delta$ indicates a more tree-like geometry (see Appendix~\ref{app:hyperbolic:hyperbolicity}). To make comparison meaningful across the different scales, we normalize the value by the task horizon and report relative hyperbolicity $\delta_{\mathrm{rel}}\in[0,1]$ instead~\citep{Khrulkov2020DeltaRelHyperbolicity, ermolov2022hyperbolic, Bdeir2024FullyHyperbolicCNNs}.

Figure~\ref{fig:analysis-fig1-delta} evaluates $\delta_{\mathrm{rel}}$ over progressively larger horizons, comparing route distances with straight-line distances between the same points (see Appendix~\ref{app:u5-case:hyperbolicity}). As the horizon grows, the maze becomes increasingly tree-like with respect to route distance, while straight-line distances show the opposite trend. In particular, Ant U5-Maze is locally close to flat and tree-like at a global scale. This transition from local Euclidean to globally tree-like geometry is characteristic of hyperbolic geometry: the hyperboloid is approximately Euclidean at scales below its curvature radius $1/\sqrt{c}$ and exhibits tree-like geometry at larger scales. Section~\ref{sec:experiments} shows that matching the critic's geometry to this structure improves performance in mazes. Importantly, this does not mean that hyperbolic representations are necessary to solve Ant U5-Maze. Euclidean networks can also learn the maze's tree-like geometry, but they may require much larger depth to do so. In contrast, hyperbolic geometry provides this tree-like structure directly as an inductive bias. Figure~\ref{fig:first_page} shows the effects of depth and geometry for Humanoid U-Maze.

\paragraph{Do embedding distances track route distance?}
Next, we study whether distant locations in the environment are associated with larger embedding distances between their representations. We evaluate every critic on the same set of 20{,}000 fixed state-goal pairs, sampled uniformly from rollouts of Euclidean and hyperbolic agents across the observed range. For each critic, we compute the Spearman rank correlation $\rho$ between embedding distance and route distance. This measures how well the critic orders goals by the distance the agent must travel to reach them. Figure~\ref{fig:analysis-fig1-correlation} shows that hyperbolic critics order goals by route distance more consistently than Euclidean critics, as seen by the significantly higher $\rho$. A high correlation alone does not indicate that the critic has learned the maze's geometry since straight-line distance also correlates with route distance. Therefore, a critic that only tracks straight-line distance has a high correlation, too. To isolate this effect, we select pairs of goals with similar straight-line distance but route distances that differ by at least 10 world units. Then, we check whether the critic assigns the larger embedding distance to the goal with the longer route. Ranking by random chance yields exactly 50\%, which a ranking by straight-line distance roughly achieves. Figure~\ref{fig:analysis-fig1-ordering} shows that hyperbolic critics order such goals more precisely than Euclidean critics. Thus, hyperbolic critics place distant locations farther apart in embedding space, even when their straight-line distance suggests otherwise.

\paragraph{Does the critic point the agent in the right direction?}
Finally, we ask whether these geometric differences change the direction in which the critic guides the agent. To that end, we compute the gradient $\nabla_{x,y} Q(s,a,g)$ of the critic with respect to the agent's position $(x,y)$ in state $s$, holding the action $a$ fixed. We evaluate this gradient on state-goal pairs that are far apart in terms of route distance but close in straight line distance, so that moving straight towards the goal leads into a wall. Hence, the agent must traverse the maze. Figure~\ref{fig:analysis-fig2-gradient-alignment} reports the cosine similarity between $\nabla_{x,y} Q$ and two directions: straight towards the goal through the wall, and the direction that minimizes route distance. Euclidean critics tend to direct the agent toward the wall rather than along the route, whereas hyperbolic critics do not exhibit this bias. The value function visualization in Figure~\ref{fig:analysis-fig2-value-func} shows the same pattern. While the Euclidean critic assigns high values across walls, the hyperbolic critic does not.

\clearpage

\section{Experiments}\label{sec:experiments}
We build on the JaxGCRL implementation of \citet{Bortkiewicz2025AcceleratingCRL} and follow the benchmark setup of \citet{Wang2025ScalingCRL}, with two changes: we replace Ant Big Maze with Ant Ball to cover a wider range of task types, and we evaluate Humanoid under a different protocol. The difference pertains to the frequency of the physics simulation, detailed in Appendix~\ref{app:envs:wang_humanoid_diff}. Appendix~\ref{app:envs} lists all used environments. We use the steps-at-goal as our performance measure, following \citet{Wang2025ScalingCRL}, and report per-environment means with 95\% bootstrap confidence intervals (CIs)~\citep{Agarwal2021RliableStatisticalPrecipice}. Both Euclidean and hyperbolic agents use a scaled ResNet encoder for the actor and the critic~\citep{Wang2025ScalingCRL}. We choose depth 16 for our main performance experiments based on our scaling results in Section~\ref{subsec:experiments:ablations}, which presents depth-scaling ablations of our models on selected environments. The hyperbolic agent adds an exponential map at the origin and a single HTC transformation~\citep{Yang2024Hypformer} after the Euclidean encoders' representation. At depth 16, this adds approximately 0.26\% to the total parameter count compared to fully Euclidean networks, but yields substantial improvements across many tasks in the full benchmark suite. We use HTC over other hyperboloid layers due to its simpler implementation, and rely on the hyperbolix library~\citep{Klein2026Hyperbolix} for JAX implementations of hyperbolic operations. We fix the curvature magnitude to $c=0.5$ (i.e., sectional curvature $-c$) for all hyperbolic agents. Our baseline is Euclidean CRL, since \citet{Wang2025ScalingCRL} show that other methods, such as HER~\citep{Andrychowicz2017HER} or CMD-1~\citep{Myers2024LearningTemporalDistances}, are not competitive.

\subsection{Main Results}\label{subsec:experiments:online}

\begin{figure}[t]
     \centering
     \includegraphics[width=\columnwidth]{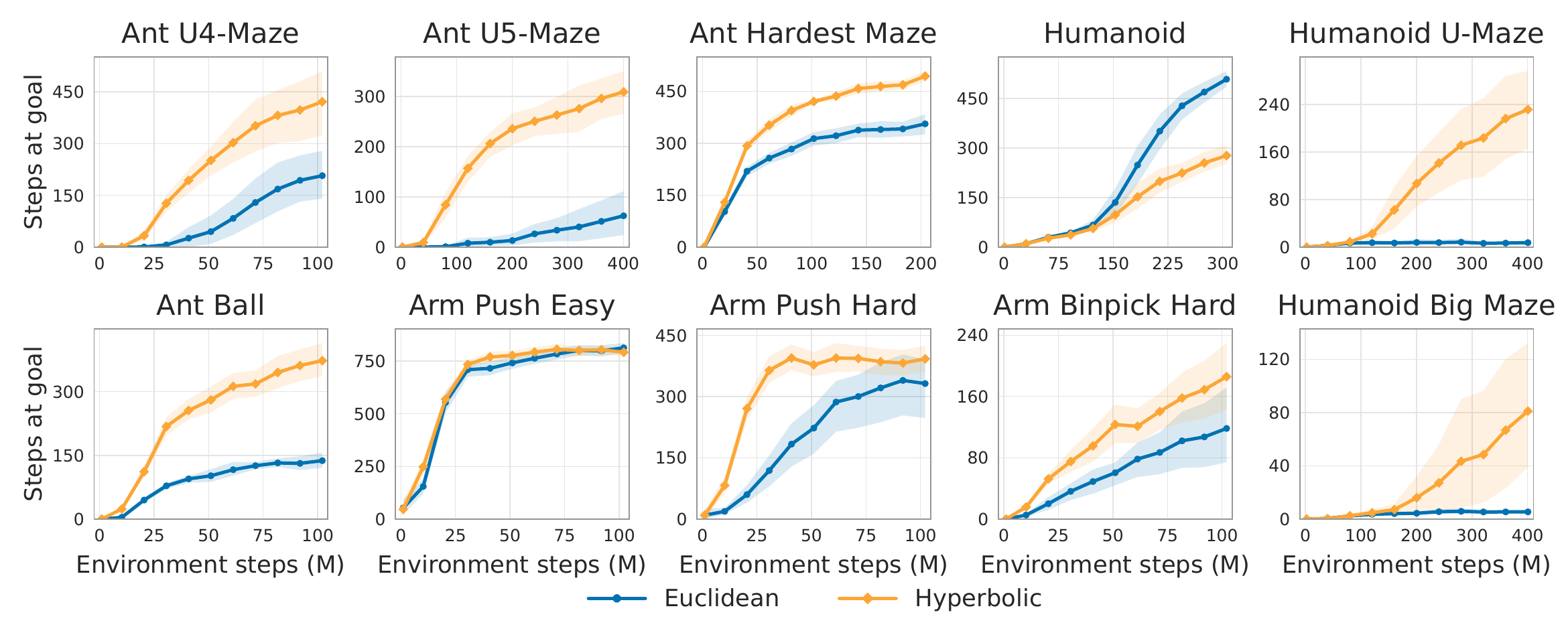}
    \caption{\textbf{Hyperbolic agents outperform Euclidean ones at fixed depth.} Hyperbolic improvements are particularly pronounced in hard navigation environments, e.g., Humanoid mazes. Humanoid is not suitable for curved representations at $c=0.5$ because route distances are straight lines. All agents use depth 16, beyond which gains flatten (Figure~\ref{fig:depth-ablation}). We report the mean across 10 seeds with 95\% bootstrapped confidence intervals.}
    \label{fig:performance_main}
    \vspace{-3mm}
\end{figure}

As Figure~\ref{fig:first_page} highlights, we find that hyperbolic agents improve the most over the baselines on the two hardest tasks in the suite, Humanoid U-Maze and Humanoid Big Maze. In these two environments, our agents substantially outperform depth 1024 and depth 512 Euclidean agents at depth 64, a fraction of the Euclidean depth. Figure~\ref{fig:performance_main} shows training curves for the full suite comprising locomotion, navigation, and manipulation tasks. The strong performance of hyperbolic agents on Ant U-Mazes connects directly to Section~\ref{sec:analysis}: evaluation in these environments uses only far-away goals, exactly the scenario where our case study predicts the strongest advantage for hyperbolic representations. In both Humanoid mazes, the hyperbolic agents make steady learning progress, while the Euclidean agent barely gets off the ground. Section~\ref{subsec:experiments:behavior_qualitative} analyzes the behavioral difference between the two agents. We discuss Humanoid and Ant Ball separately because both environments are open arenas without walls. First, Humanoid's difficulty lies in locomotion, and every route to the goal is a straight line in the plane. Without walls, route distance generally equals straight-line distance, which Section~\ref{sec:analysis} finds far less tree-like than route distance in mazes. We therefore expect no hyperbolic advantage on Humanoid. The observed gap comes from our default curvature $c = 0.5$, which imposes a strongly curved geometry on a \quoteeng{flat} environment. A lower curvature closes the gap, as our ablations in Section~\ref{subsec:experiments:ablations} show. Ant Ball uses the same arena layout as Humanoid but requires pushing a ball to the goal. Surprisingly, the hyperbolic agent wins here. Visual inspection of the learned policies in Figure~\ref{app:fig:ant_ball_behavior} reveals that hyperbolic agents reposition themselves around the ball more often and push harder. This repositioning requires curved paths even in an open arena, and we hypothesize that this is why hyperbolic representations help on this task.

\subsection{Behavioral Evidence of Invalid Shortcuts in Euclidean Agents}\label{subsec:experiments:behavior_qualitative}

\begin{wrapfigure}{r}{0.5\linewidth}
    \centering
        \vspace{-3mm}
        \includegraphics[width=\linewidth]{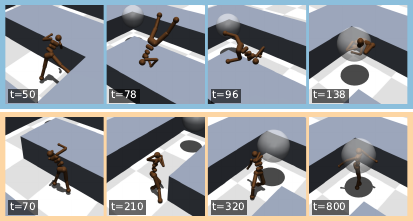}
        \caption{\textbf{Behavioral evidence of straight-line bias}. Similar to \citet{Wang2025ScalingCRL}, we observe Euclidean agents trying to vault themselves over walls instead of learning to navigate the maze. In contrast, hyperbolic agents follow the corridors, learn locomotion behavior earlier, and achieve higher steps at goal.} 
        \label{fig:online_qualitative}
        \vspace{-3mm}
\end{wrapfigure}

The straight-line bias discussed in Section~\ref{sec:analysis} is also evident in the behavior of Euclidean agents. Figure~\ref{fig:online_qualitative} shows a representative successful trajectory of each agent on Humanoid U-Maze for a distant goal that requires navigating around the wall. The Euclidean agent takes a straight-line shortcut: it runs at the wall, vaults over it, lands in the goal region, and falls, which ends the episode. The hyperbolic agent instead walks around the corner and stands at the goal. \citet{Wang2025ScalingCRL} report a similar vault for depth 256 Euclidean agents and present it as a new capability. Our depth 16 Euclidean agent also vaults, but falls on landing. We consider this vault undesirable because it achieves the goal without learning the locomotion behavior required to stay on it. Consistent with this, the depth 1024 Euclidean agents of \citet{Wang2025ScalingCRL} spend around 100 steps at the goal on Humanoid U-Maze, whereas our depth 64 hyperbolic agent stays for 240. One consistent difference between the agents across all seeds is locomotion behavior. Euclidean agents typically fall within 90 steps and barely manage to stay at the goal, even when the goal is in the same corridor as the starting state. Hyperbolic agents walk to these goals and stay at them in most episodes. Goals behind the inner wall remain difficult for both agents. Scaling Euclidean networks up to depth 64 still yields agents that rarely stay at goals, consistent with the collapse reported by \citet{Wang2025ScalingCRL} at depth 64. We hypothesize that the straight-line bias of Section~\ref{sec:analysis} contributes to the Euclidean agents' failure to learn a stable gait at lower depths because they try to take the shortcut over the wall.

\subsection{Hyperbolic Representations Improve Stitching}\label{subsec:experiments:stitching}

\begin{wrapfigure}{r}{0.4\linewidth}
    \centering
        \includegraphics[width=\linewidth]{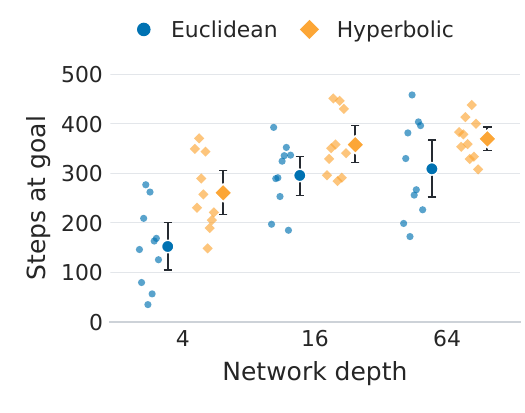}
        \caption{\textbf{Stitching on Ant U-Maze}. Markers show the mean steps at goal with 95\% bootstrap CIs.} 
        \label{fig:online_stitching}
        \vspace{-3mm}
\end{wrapfigure}

Hyperbolic representations also improve experience stitching, a form of generalization intrinsic to deep RL~\citep{Bortkiewicz2025Stitching}. We follow the protocol of \citet{Wang2025ScalingCRL} on Ant U-Maze, where agents train only on start-goal pairs at most three units apart and are evaluated on the three goals on the other side of the connecting corridor. This means that every evaluation pair requires combining short training trajectories. Figure~\ref{fig:online_stitching} shows that hyperbolic agents stay on average longer at the goal than Euclidean agents at every depth, with the largest gap at depth 4. At depth 16, hyperbolic agents already exceed the mean of depth 64 Euclidean agents. Beyond depth 16, gains in mean performance flatten for both geometries, but hyperbolic agents become more consistent across seeds as depth increases to 64, narrowing the confidence intervals. Section~\ref{sec:analysis} suggests why hyperbolic agents stitch better: the evaluation goals lie close to the start in straight-line distance but far apart in the route distance. This aligns with our previous analysis of Figure~\ref{fig:analysis-fig1-ordering}, showing that hyperbolic critics order such goals by route distance more reliably than Euclidean critics. Because these long-range start--goal pairs never appear in training, the critic must infer their distances from shorter trajectories. Hyperbolic geometry provides the inductive bias for this long-range structure, whereas Euclidean critics must learn it through depth. This explains the large gap at depth 4.

\subsection{Performance Sensitivity across Curvature and Network Depth}\label{subsec:experiments:ablations}

\begin{wrapfigure}{r}{0.4\linewidth}
    \centering
         \vspace{-5mm}
        \includegraphics[width=\linewidth]{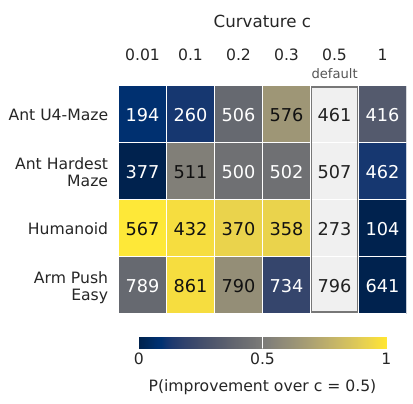}
        \caption{\textbf{Effect of curvature.} The best curvature setting depends on the environment, with Humanoid preferring low curvature, and Ant environments preferring higher curvature. Mean over the final 10 evaluation epochs over five seeds.} 
        \label{fig:curvature_ablation}
        \vspace{-3mm}
\end{wrapfigure}

The exact geometry of the hyperboloid depends on its sectional curvature $-c$, where $c \rightarrow 0$ implies Euclidean geometry. To analyze how curvature affects performance, we run a sensitivity analysis over the magnitudes $c \in \{0.01, 0.1, 0.2, 0.3, 0.5, 1\}$ across four tasks, each with five seeds. Figure~\ref{fig:curvature_ablation} compares every setting against our default $c = 0.5$ by the probability that a run at that curvature outperforms a run at $c = 0.5$~\citep{Agarwal2021RliableStatisticalPrecipice}. Figure~\ref{app:fig:curvature_ablation} shows additional results and compares against Euclidean agents. We find that intermediate curvatures work best on mazes. On Ant U4-Maze and Ant Hardest Maze, reducing $c$ to 0.01 drops performance to the Euclidean level, and on Ant U4-Maze, $c = 0.1$ also falls below the default curvature. At the other end, $c = 1$ performs worse than $c = 0.5$ on all four tasks. Arm Push Easy prefers a lower curvature, with $c=0.1$ yielding the best results. Humanoid admits a different geometry, which explains our $c=0.5$ results in Section~\ref{subsec:experiments:online}. Here, the performance of hyperbolic agents increases monotonically as the curvature magnitude $c$ decreases. Our flattest setting, $c = 0.01$, matches the Euclidean performance. Unsurprisingly, open arenas favor flat geometries, where route distance equals straight-line distance. In mazes, where routes bend around walls, the same flat geometry removes the hyperbolic advantage.

\begin{figure*}[t]
  \centering
      \begingroup
    \definecolor{depthEuclidean}{HTML}{0072B2}%
    \definecolor{depthHyperbolic}{HTML}{FCA636}%
    \small
    \tikz[baseline=-0.5ex]{%
      \draw[depthEuclidean,line width=0.95pt] (0,0) -- (0.52,0);
      \fill[depthEuclidean] (0.26,0) circle[radius=2pt];
    }%
    \hspace{0.35em}Euclidean\hspace{1.8em}%
    \tikz[baseline=-0.5ex]{%
      \draw[depthHyperbolic,line width=0.95pt] (0,0) -- (0.52,0);
      \fill[depthHyperbolic] (0.26,0.105) -- (0.365,0) -- (0.26,-0.105) -- (0.155,0) -- cycle;
    }%
    \hspace{0.35em}Hyperbolic
    \endgroup
  \par\smallskip
  
  \captionsetup[subfigure]{position=top,font=small,justification=centering,singlelinecheck=true,skip=2pt}
  \begin{subfigure}[t]{0.245\textwidth}
    \centering
    \caption{Ant U4-Maze}
    \label{fig:depth-ant-u4}
    \includegraphics[width=\linewidth]{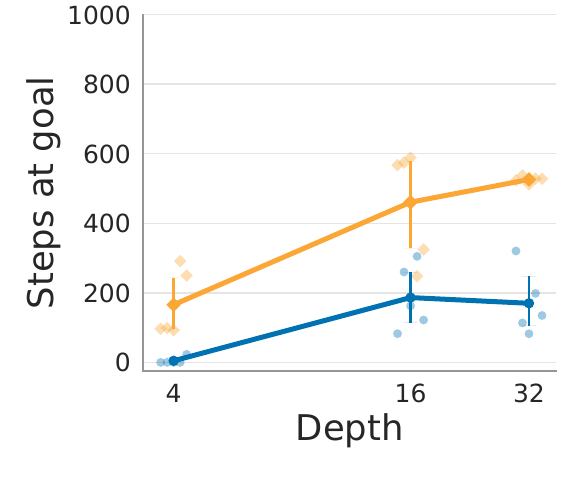}
  \end{subfigure}\hfill%
  \begin{subfigure}[t]{0.245\textwidth}
    \centering
    \caption{Ant Ball}
    \label{fig:depth-ant-ball}
    \includegraphics[width=\linewidth]{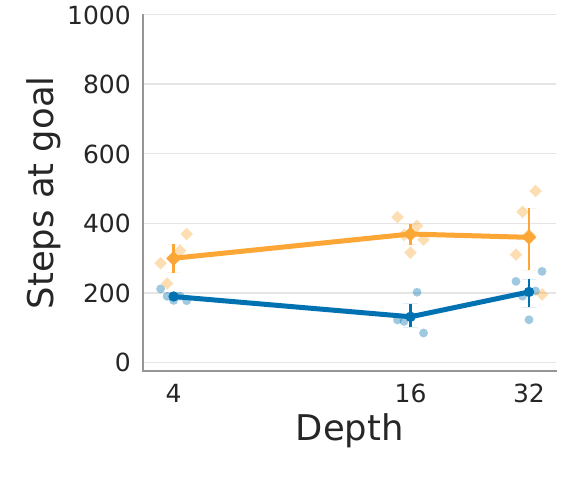}
  \end{subfigure}\hfill%
  \begin{subfigure}[t]{0.245\textwidth}
    \centering
    \caption{Arm Push Easy}
    \label{fig:depth-arm-push}
    \includegraphics[width=\linewidth]{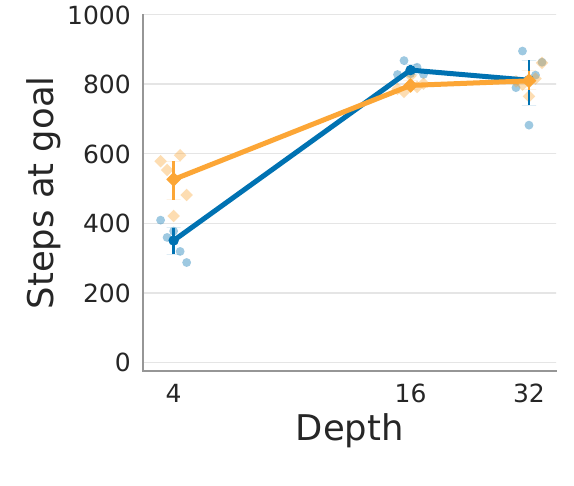}
  \end{subfigure}\hfill%
  \begin{subfigure}[t]{0.245\textwidth}
    \centering
    \caption{Humanoid}
    \label{fig:depth-humanoid}
    \includegraphics[width=\linewidth]{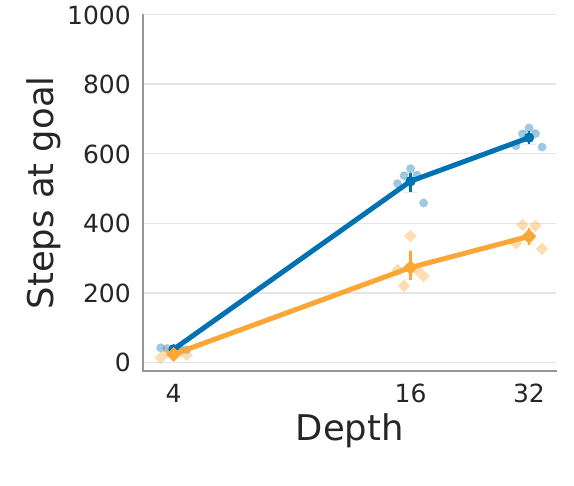}
  \end{subfigure}
  \caption{\textbf{Effect of actor and critic depth scaling.} Hyperbolic agents generally perform better as the Euclidean encoder grows deeper, although the size of the gain varies across environments. We report the mean over five seeds with 95\% bootstrap confidence intervals.}
  \label{fig:depth-ablation}
  \vspace{-5mm}
\end{figure*}

Hyperbolic agents benefit from a deeper Euclidean encoder on three of the four tasks and maintain their performance on Ant Ball (Figure~\ref{fig:depth-ablation}). On Humanoid, hyperbolic agents improve with depth like Euclidean agents, but remain well below them from depth 16 on. The curvature ablation in Figure~\ref{fig:curvature_ablation} attributes this deficit to our default curvature magnitude of $c = 0.5$. On Ant U4-Maze, the hyperbolic advantage between the two methods widens with depth, and a 4-layer hyperbolic agent already matches Euclidean agents with 16 or 32 layers. On Arm Push Easy, depth instead closes the gap, and both methods perform alike from 16 layers on. For both methods, the gains flatten as the network grows. Beyond 16 layers, Euclidean agents improve only on Humanoid. We therefore use 16 Euclidean layers for the main results in Figure~\ref{fig:performance_main}.

\clearpage

\subsection{Offline Experiments}\label{sec:offline}

\begin{figure}[t]
\centering
  \begingroup
  \definecolor{depthEuclidean}{HTML}{0072B2}%
  \definecolor{depthHyperbolic}{HTML}{FCA636}%
  \small
  \tikz[baseline=-0.5ex]{%
    \fill[depthEuclidean] (0.26,0) circle[radius=2pt];
  }%
  \hspace{0.35em}Euclidean\hspace{1.8em}%
  \tikz[baseline=-0.5ex]{%
    \fill[depthHyperbolic] (0.26,0.105) -- (0.365,0)
      -- (0.26,-0.105) -- (0.155,0) -- cycle;
  }%
  \hspace{0.35em}Hyperbolic
  \endgroup
  \par\smallskip

\captionsetup[subfigure]{font=footnotesize,position=top,skip=1pt,justification=centering,singlelinecheck=false}
\begin{subfigure}[t]{0.444866\textwidth}
\centering
\caption{AntMaze}\label{fig:ogbench-antmaze}
\includegraphics[width=\linewidth]{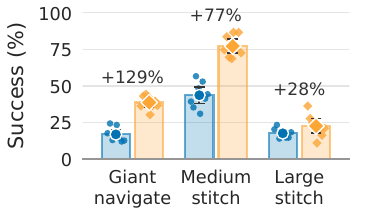}
\end{subfigure}%
\begin{subfigure}[t]{0.555134\textwidth}
\centering
\caption{HumanoidMaze}\label{fig:ogbench-humanoidmaze}
\includegraphics[width=\linewidth]{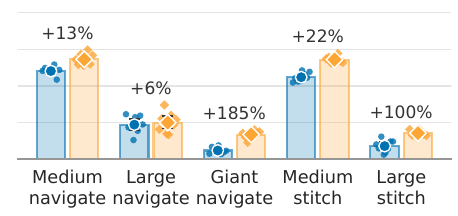}
\end{subfigure}
\caption{\textbf{Offline results on OGBench}. The hyperbolic agent reaches a higher mean success rate on all eight tasks, with the largest relative gains on the giant mazes. Percentages give the relative improvement of the hyperbolic mean over the Euclidean mean. Bars show the mean over 8 seeds with 95\% bootstrap CIs.}
\label{fig:offline_results}
\vspace{-5mm}
\end{figure}

Increasing network depth does not reliably improve offline GCRL. As \citet[Figure 18]{Wang2025ScalingCRL} scale the depth of Euclidean CRL from 4 to 64 layers, performance drops sharply on antmaze-giant-navigate and humanoidmaze-medium-navigate. On antmaze-medium-stitch, they report moderate gains. To test whether hyperbolic agents fail in the same way, we build on the CRL implementation of OGBench~\citep{Park2025OGBench} and adapt our online agent accordingly, see Appendix~\ref{app:arch_hparams:hparams_offline} for all changes. Following \citet{Park2025OGBench}, we use the binary success rate as an evaluation metric rather than steps at goal. Figure~\ref{fig:offline_results} shows that the hyperbolic agent reaches a higher mean success rate than the Euclidean agent on all eight tasks. The relative gains are largest on the two giant mazes, although neither agent reliably solves them. On navigation, the hyperbolic agent improves clearly on three of the four tasks, including antmaze-giant-navigate and humanoidmaze-medium-navigate, the two tasks on which deeper Euclidean agents fail \citep[Figure 18]{Wang2025ScalingCRL}. The hyperbolic agent also shows clear gains on three of the four stitching tasks, with overlapping confidence intervals only for antmaze-large-stitch. This matches our online stitching results in Section~\ref{subsec:experiments:stitching}. The largest absolute gain is on antmaze-medium-stitch, where our hyperbolic agent exceeds 75\% success, above the roughly 70\% that the depth 64 residual Euclidean agent reaches in~\citet[Figure 18]{Wang2025ScalingCRL}. Across the three OGBench tasks that \citet{Wang2025ScalingCRL} evaluate, depth helps on the stitching task and hurts on both navigation tasks. Hyperbolic geometry improves all three.

\section{Conclusion}\label{sec:conclusion}

We introduce hyperbolic CRL, a drop-in modification to CRL that replaces the Euclidean distance in the critic with a hyperbolic geodesic distance, providing the agent with an inductive bias towards tree-like geometry at a small computational cost. With this change, depth 64 hyperbolic agents outperform 1024-layer Euclidean agents on hard humanoid maze tasks. The gains follow our geometric analysis in Section~\ref{sec:analysis}. They are largest where walls cause the route distance to deviate from the straight-line distance, and disappear on Humanoid, where route distances coincide with straight distances. More broadly, our results suggest that matching the representation geometry to the environment structure can alleviate much of the network depth needed for long-horizon goal reaching.

\paragraph{Limitations and future work.}  The results on Humanoid and Arm Push show that hyperbolic agents can match Euclidean performance, but with substantially lower curvature than our default of $c=0.5$. This suggests that learning an environment-dependent curvature value improves performance. However, we found that dynamically setting or learning the curvature is challenging due to the noise in deep RL training and variability in agent performance. Second, geodesic distance is symmetric, so the critic cannot represent one-way transitions. We believe that combining hyperbolic geometry with quasimetric structure is a natural extension for solving this. 

\newpage

\subsection*{AI use statement}

In this work, we used generative AI tools for literature search, coding assistance, running experiments, writing assistance, and figure generation. In the literature search, we used AI tools to find and summarize relevant work, which was then manually checked for relevance. For code, we used AI tools for repetitive tasks, such as porting code from one deep learning framework to another and generating matplotlib figures. We reviewed our code, tested it, and verified that our baselines reproduce the results reported in the literature. For the paper, generative AI was used as a tool to iterate on our writing. An example use case is paragraph shortening with \quoteeng{Improve the clarity and conciseness of this paragraph.} The idea to use hyperbolic representations in self-supervised RL, our analysis case study, and our ablation studies were generated and designed by us. We take responsibility for the final content of this work, including text, claims, or artifacts produced with the aid of generative AI.

\subsection*{Ethics statement}

This work studies the representation geometry of self-supervised goal-conditioned RL agents and raises no direct ethical concerns on its own. All of our experiments are conducted on simulated locomotion and manipulation benchmarks and involve no human data. We cannot rule out downstream misuse because agents that learn to reach arbitrary goals without hand-designed rewards are a building block for autonomous robotic systems. However, such misuse would require substantial extensions beyond the simulated setting evaluated here.

\subsection*{Reproducibility statement}

Our code is included in the submission and will be made publicly available upon acceptance. Our environment setup is in Appendix~\ref{app:envs}. Notably, our evaluation protocol differs from \citet{Wang2025ScalingCRL} on Humanoid, which we discuss in Appendix~\ref{app:envs:wang_humanoid_diff}. We provide a schematic of our architecture in Appendix~\ref{app:arch_hparams}. Appendices~\ref{app:arch_hparams:hparams_online} and~\ref{app:arch_hparams:hparams_offline} contain hyperparameters for our online and offline experiments, respectively.  Lastly, wall-clock times for our architecture on an H100 GPU are listed in Appendix~\ref{app:wall_clock}.

\subsubsection*{Acknowledgments}
This work has been funded in parts by the Vienna Science and Technology Fund (WWTF) [10.47379/ICT20058]. We acknowledge EuroHPC JU for awarding the projects EHPC-DEV-2026D07-077 and EHPC-DEV-2026D08-151 access to the Luxembourg national supercomputer (MeluXina), the Portuguese supercomputer Deucalion, and the Slovenian HPC system Vega at the Institute of Information Science. The authors gratefully acknowledge LuxProvide, the Minho Advanced Computing Center (MACC), and the Institute of Information Science (IZUM) for their expert support. Without Nikolaus Süß' tireless work maintaining our research group's computing infrastructure, this work would not have been possible. We are deeply grateful. Lastly, we want to thank the open source community, particularly the developers of JAX~\citep{Bradbury2018JAX}, Flax~\citep{Heek2020Flax}, and Matplotlib~\citep{Hunter2007Matplotlib}.

\bibliography{refs_rl,refs_hyp,refs_other}
\bibliographystyle{iclr2027_conference}

\input{yAppendix}

\end{document}

%% file: math_commands.tex
\usepackage{amsmath,amsfonts,bm}

\def\eqref#1{equation~\ref{#1}}

\def\1{\bm{1}}

\DeclareMathAlphabet{\mathsfit}{\encodingdefault}{\sfdefault}{m}{sl}
\SetMathAlphabet{\mathsfit}{bold}{\encodingdefault}{\sfdefault}{bx}{n}

%% file: yAppendix.tex
\appendix
\section{Appendix}

\WFclear

\section{Extended Background on Hyperbolic Geometry}\label{app:hyperbolic}

\subsection{Hyperbolic Space Operations}\label{app:hyperbolic:hyp_ops}
Following the terminology of special relativity, we refer to the first coordinate $x_0$ of a point $\bm{x} \in \hyperboloid$ as its \emph{time component} and to the remaining coordinates $\bm{x}_{1:n} = (x_1, \dots, x_n)$ as its \emph{space component}.

\subsubsection{Exponential Map}\label{app:hyperbolic:expmap}
The \textit{exponential map} $\expmap{x}:\htangent{x}\rightarrow\hyperboloid$ at a point $\bm{x}\in\hyperboloid$ maps a tangent vector $\bm{v}\in\htangent{x}=\{\bm{v} : \minner{\bm{v}}{\bm{x}}=0\}$ to the manifold such that the curve $t\mapsto\expmap{x}(t\bm{v})$ is the geodesic starting at $\bm{x}$ with initial velocity $\bm{v}$. Since the hyperboloid is geodesically complete, $\expmap{x}$ is defined on the entire tangent space and given by
\[\expmap{x}(\bm{v}) = \cosh\left(\sqrt{c}\,\mnorm{\bm{v}}\right)\bm{x} + \sinh\left(\sqrt{c}\,\mnorm{\bm{v}}\right)\frac{\bm{v}}{\sqrt{c}\,\mnorm{\bm{v}}}.\]
We use the exponential map at the origin to project Euclidean input features onto the manifold. Identifying $\bm{u}\in\mathbb{R}^d$ with the tangent vector $\bm{v}=(0,\bm{u})\in\htangent{\horigin}$, where $\horigin=(1/\sqrt{c},\bm{0})$, yields
\[\expmap{\horigin}(\bm{v}) = \left(\frac{1}{\sqrt{c}}\cosh(\sqrt{c}\,\|\bm{u}\|),\ \frac{\sinh(\sqrt{c}\,\|\bm{u}\|)}{\sqrt{c}\,\|\bm{u}\|}\,\bm{u}\right).\]

\subsubsection{Distance}\label{app:hyperbolic:hyp_dist}
For any two points $\bm{x}, \bm{y} \in \hyperboloid$, the \textit{geodesic distance} $d_{\hyperboloid}(\bm{x},\bm{y})$ is the length of the shortest path on the manifold connecting them. In the hyperboloid model, it depends only on their Lorentzian inner product:
\[d_{\hyperboloid}(\bm{x},\bm{y}) = \frac{1}{\sqrt{c}}\,\text{arcosh}\left(-c\,\minner{\bm{x}}{\bm{y}}\right).\]

\subsubsection{Hyperbolic Linear Layers}\label{app:hyperbolic:hyp_linear_layers}
We use the fully connected hyperbolic linear transformation of \citet{Yang2024Hypformer}, which operates directly on the hyperboloid without mapping to and from a tangent space. Given $\bm{x}\in\hyperboloid$, a weight matrix $\bm{W}\in\mathbb{R}^{(n+1)\times n'}$, and a bias $\bm{b}\in\mathbb{R}^{n'}$, it first applies the Euclidean affine map $f(\bm{x}) = \bm{W}^\top\bm{x}+\bm{b}$ to the full input, including its time-like component. The result vector is then treated as the space-like components of the output. The time-like component is determined by the hyperboloid constraint $\minner{\bm{y}}{\bm{y}}=-1/c$:
\[\bm{y} = \left(\sqrt{\|f(\bm{x})\|^2 + \frac{1}{c}},\ f(\bm{x})\right)\in\mathbb{H}^{d'}_{c}.\]

\subsection{Delta-hyperbolicity}\label{app:hyperbolic:hyperbolicity}
\begin{figure}[h!]
  \centering
  {\sansmath\sffamily
    \resizebox{0.6\textwidth}{!}{%
      \input{tikz/gromov_product.tikz}%
    }%
  }
  \caption{Geometric interpretation of the Gromov product. \emph{Left:} three points $w$, $x$, $z$ of a metric space, with their pairwise distances. \emph{Right:} the associated comparison tripod, i.e. the unique metric tree with leaves $w$, $x$, $z$ joined at a branching point $p$ in which the path between any two leaves is of the same length as the corresponding distance on the left. The quantity $(x,z)_w = \tfrac12\bigl(d(w,x) + d(w,z) - d(x,z)\bigr)$ is the length of the initial segment that shortest paths from $w$ to $x$ and from $w$ to $z$ share in the tripod before they separate at $p$.}
  \label{fig:gromov_product}
\end{figure}
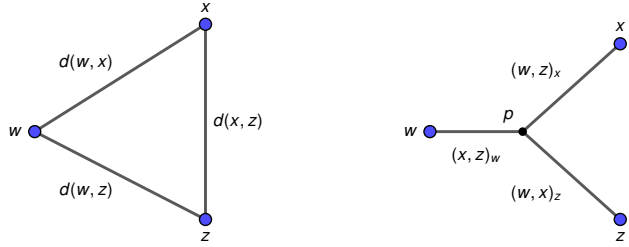

To quantify how tree-like an embedding or environment is, we use Gromov $\delta$-hyperbolicity, formally defined in terms of the Gromov product. For a metric space $(X,d)$ and points $w,x,z\in X$, the Gromov product (Figure~\ref{fig:gromov_product}) of $x$ and $z$ with respect to $w$ is
\[(x,z)_{w}={\frac {1}{2}}{\big (}d(w,x)+d(w,z)-d(x,z){\big )}.\]

A metric space is called $\delta$-hyperbolic if, for all $x,y,z,w\in X$,
\[(x,z)_{w}\geq \min \left\{(x,y)_{w},(y,z)_{w}\right\}-\delta.\]

The hyperbolicity constant of the space is the smallest $\delta \ge 0$ for which this inequality holds for all quadruples of points. Under this framework, Gromov-hyperbolic spaces can be viewed as metric spaces that are tree-like up to a bounded additive distortion. In particular, metric trees are $0$-hyperbolic, while Euclidean spaces $\mathbb{R}^n$ with $n\geq2$ are not $\delta$-hyperbolic for any finite $\delta$.

While $\delta$ provides an absolute measure of tree-likeness, its value cannot be compared directly across finite metric spaces of different diameters. To account for this the hyperbolicity constant is typically normalized by the diameter of the metric space
\[\delta_{\mathrm{rel}} = \frac{2\,\delta(X)}{\operatorname{diam}(X)}\;\in[0,1]\]
to provide a scale-invariant quantity. Values close to $0$ indicate that deviations from tree-like geometry are small relative to the diameter of the space, and values close to $1$ indicate the largest deviation possible at that scale.

\clearpage

\section{Environment Setup}\label{app:envs}

\subsection{Online Environments}\label{app:online_envs}

\begin{figure}
    \centering
        \includegraphics[width=\linewidth]{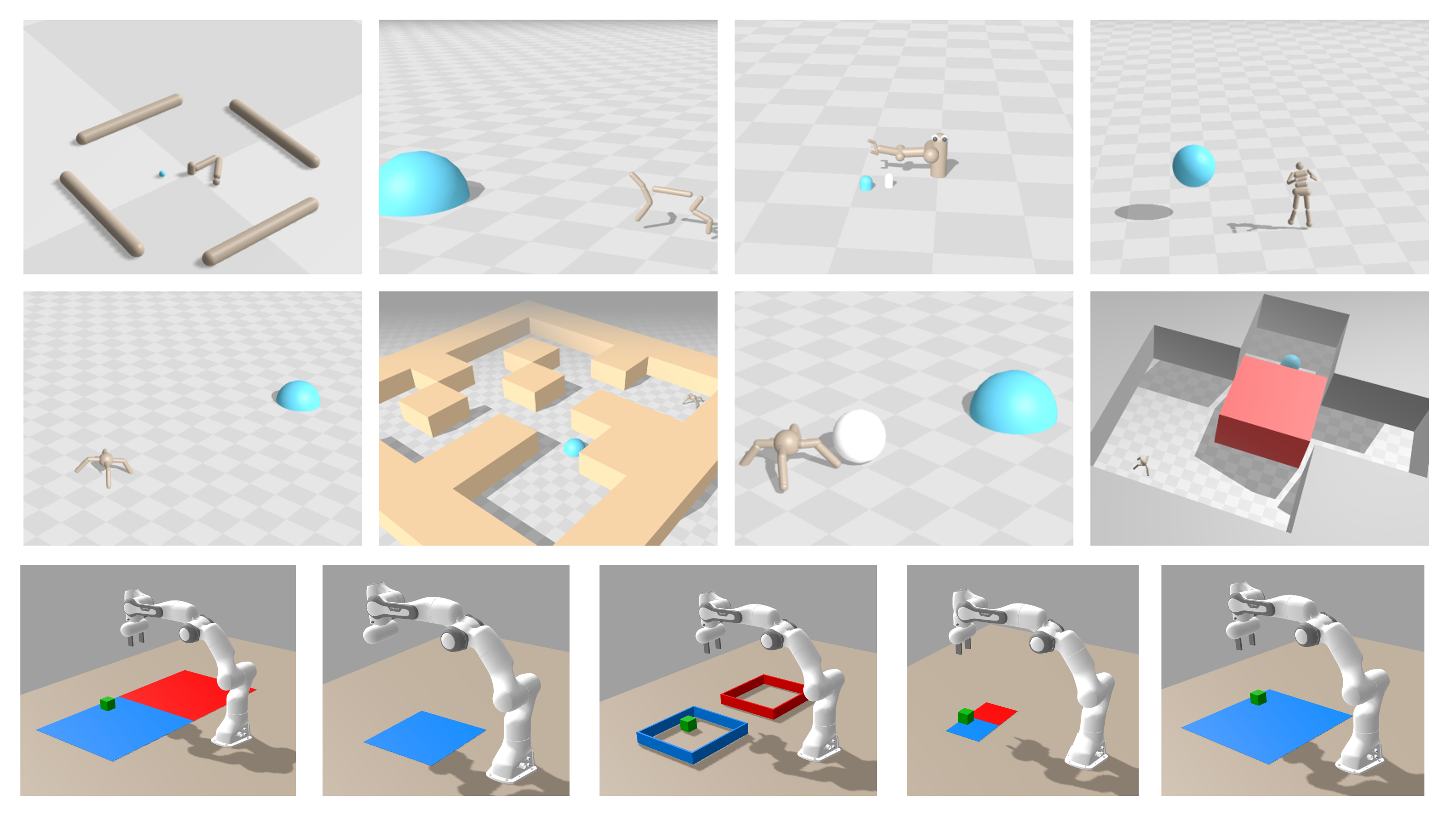}
        \caption{\textbf{Example Environments.} We use the environments implemented in JaxGCRL~\citep{Bortkiewicz2025AcceleratingCRL}. Figure taken from \citet{Bortkiewicz2025AcceleratingCRL}.} 
        \label{app:fig:envs}
        \vspace{-3mm}
\end{figure}

We use the online environment implementations of \citet{Bortkiewicz2025AcceleratingCRL}. The full list of our 10 environments is: \texttt{arm\_binpick\_hard}, \texttt{arm\_push\_easy}, \texttt{arm\_push\_hard}, \texttt{humanoid}, \texttt{humanoid\_big\_maze}, \texttt{humanoid\_u\_maze}, \texttt{ant\_ball}, \texttt{ant\_hardest\_maze},  \texttt{ant\_u4\_maze}, \texttt{ant\_u5\_maze}.

\subsection{Details on the Ant U5 Maze Case Study}\label{app:u5-case}

\paragraph{Measuring hyperbolicity.}\label{app:u5-case:hyperbolicity}
We quantify how tree-like the geometry of the Ant U5 maze is by evaluating its relative $\delta$-hyperbolicity (Appendix~\ref{app:hyperbolic:hyperbolicity}) as a function of the task horizon. Starting from its local structure (short horizons, nearby goals) we iteratively expand the environment to its global one (long horizons, far away goals). For comparison, we track both \emph{straight-line distances} (Euclidean, ignoring walls) and \emph{route distances} (shortest path inside the maze) across the same set of points.

To do so, we discretize the environment into a grid with a spacing of $0.5$ world units, define route distances as the shortest paths between neighboring cells, and sample $300$ grid points uniformly at random. From these we draw $200{,}000$ quadruples of four distinct points uniformly at random. For a horizon $R$, let $X_R\subseteq X$ denote the subspace of sampled points with their quadruple lying entirely in $X_R$. We compute their relative $\delta$-hyperbolicity as
\[\delta_{\mathrm{rel}}(R):=\frac{2\,\delta(X_R)}{\operatorname{diam}(X_R)}.\]
A decreasing $\delta_{\mathrm{rel}}(R)$ therefore means that the environment becomes increasingly tree-like at larger task horizons. Straight-line distances are evaluated on the same subspaces and quadruples, normalized by the straight-line diameter of $X_R$.

As a reference, the maximum relative defect any four points in the Euclidean plane can attain is $2-\sqrt2\approx0.586$. This theoretical ceiling is reached by the vertices of a square.

\paragraph{Embedding distance versus route distance correlations.}\label{app:u5-case:correlations}
We combine 400 recorded states from each of the ten Euclidean and ten hyperbolic runs into a shared dataset. All critics evaluate the same 20{,}000 state-goal pairs, with
2{,}000 pairs per equal-width route-distance bin over approximately 0--51 world units. Each state's rollout action is fixed across critics and goals. For each critic $k$, we compute the Spearman correlation between route distance $\ell_i$ and embedding distance $e_{k,i}=d(\phi_k(s_i,a_i),\psi_k(g_i))$ across the same $N=20{,}000$ state--goal pairs.
Let $R_i$ and $S_{k,i}$ denote the ranks of $\ell_i$ and $e_{k,i}$, respectively, assigning average ranks to ties. With $\bar R$ and $\bar S_k$ denoting their mean ranks
\begin{equation}\label{eq:spearman}
    \rho_k =
    \frac{\sum_{i=1}^{N}(R_i-\bar R)(S_{k,i}-\bar S_k)}
    {\sqrt{\sum_{i=1}^{N}(R_i-\bar R)^2
           \sum_{i=1}^{N}(S_{k,i}-\bar S_k)^2}}.
\end{equation}
For each method, we report the mean correlation across 10 training seeds, with 95\% bootstrap CIs~\citep{Agarwal2021RliableStatisticalPrecipice}. The probability of improvement is computed according to \citet{Agarwal2021RliableStatisticalPrecipice}.

\paragraph{Route ordering.}\label{app:u5-case:ordering}
We draw 300{,}000 candidate comparisons with replacement. For each comparison, we select one of the 20{,}000 state--goal pairs uniformly, then select a second uniformly from the same 1-world-unit straight-line-distance bin. We retain comparisons whose route lengths differ by at least 10 world units, yielding 81{,}007 comparisons involving 13{,}666 distinct state--goal pairs. All critics are evaluated on these same comparisons. We report the percentage of comparisons in which the longer route has the larger embedding distance, assigning half credit when the embedding distances are equal. Correlation and ordering results report ten-seed means and 95\% bootstrap CIs from 50{,}000 resamples~\citep{Agarwal2021RliableStatisticalPrecipice}.

\paragraph{Gradient alignment.}\label{app:u5-case:gradient-alignment} We collect 150 rollout states per run, with test goals comprising the maze’s designated goal locations and locations visited during the collected rollouts. Actions come from each state's source policy under its original goal and remain fixed across critics and probe goals. We differentiate the critic value (negative embedding distance) with respect to torso position, holding all other inputs fixed. Cosine similarity is measured with the straight-line direction toward the goal and the direction toward the free neighboring grid node with the smallest remaining route distance. We retain state-goal pairs with a route distance of at least 24 world units and a straight-line distance below 40\% of the route distance. Additionally, we require a finite and nonzero gradient. We exclude examples where no adjacent node reduces the remaining route distance. Within each seed, we average over state-goal pairs in the $[24,48)$ and $[48,\infty)$ route-distance bins containing at least 30 valid samples. We report ten-seed means and 95\% bootstrap CIs~\citep{Agarwal2021RliableStatisticalPrecipice}.

\subsection{Humanoid Evaluation Protocol}\label{app:envs:wang_humanoid_diff}

Our Humanoid results are not directly comparable to Figure~1 of \citet{Wang2025ScalingCRL}. The discrepancy stems from how Brax stores the simulation timestep. The JaxGCRL Humanoid environment overrides the physics sub-step of its MuJoCo model from $0.003$\,s to $0.0015$\,s~\citep{Bortkiewicz2025AcceleratingCRL}. Up to version 0.10.3, Brax keeps a separate copy of the timestep that is not overridden, which is invisible from the environment code. \citet{Wang2025ScalingCRL} use Brax~0.10.1 for all their plots. This means their environment simulates at $0.003$\,s, and each environment step spans $0.03$\,s. Starting from version 0.10.4, Brax reads a single-timestep field. Consequently, the override takes effect, and each environment step spans $0.015$\,s. The version-dependent learning curves they report are consistent with the timestep rather than with other changes in the packages. Humanoid is the only affected task because the other environments, including Humanoid U-Maze and Humanoid Big Maze, respect the timestep setting via this override. We run Humanoid at the $0.0015$\,s substep that we believe \citet{Wang2025ScalingCRL} intended. This corresponds to the left panel of their Figure~20, and our Euclidean baseline is consistent with that curve.

\clearpage

\section{Architecture \& Hyperparameters}\label{app:arch_hparams}

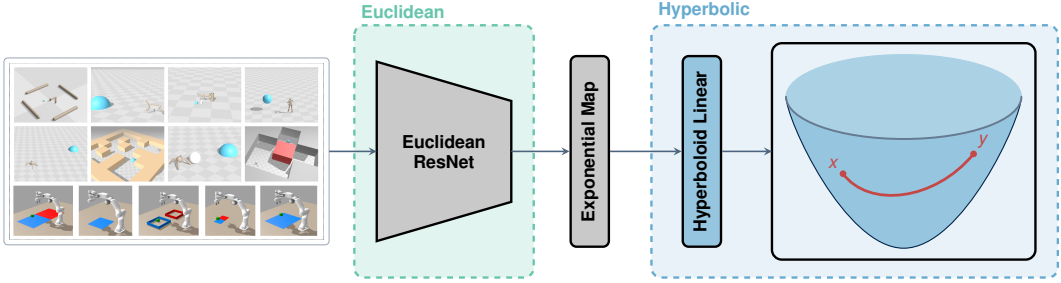
\begin{figure}[h!]
  \centering
  {\sansmath\sffamily
    \resizebox{\textwidth}{!}{%
      \input{tikz/model_architecture.tikz}%
    }%
  }
  \caption{A Euclidean encoder maps observations to feature vectors, which are then viewed as tangent vectors at the origin of the hyperboloid, mapped to the manifold via the exponential map followed by one hyperbolic linear layer (HTC, \citet{Yang2024Hypformer}). The similarity for the online goal-oriented contrastive learning is the negative geodesic distance between any two given points.}
  \label{fig:model-architecture}
\end{figure}

\subsection{Online CRL Hyperparameters}\label{app:arch_hparams:hparams_online}

  \begin{table}[h]
  \caption{\textbf{Online hyperparameters.} Unshaded rows are \citet[Table~7]{Wang2025ScalingCRL} and are shared by the
  Euclidean and the hybrid agents. \colorbox{blue!10}{Blue} rows exist only for hyperbolic agents. \colorbox{red!10}{Red} rows differ from the released code's defaults. Total number of steps per task: ant\_u4\_maze, ant\_ball, arm\_push\_easy, arm\_push\_hard, arm\_binpick\_hard 100M; ant\_hardest\_maze 200M; humanoid 300M; ant\_u5\_maze,
  humanoid\_u\_maze, humanoid\_big\_maze 400M.}
  \label{tab:hparams_online}
  \centering
  \begin{tabular}{ll}
  \toprule
  Hyperparameter & Value \\
  \midrule
  num\_timesteps & 100M--400M (varying across tasks) \\
  update-to-data (UTD) ratio & 1:40 \\
  max\_replay\_size & 10{,}000 \\
  min\_replay\_size & 1{,}000 \\
  episode\_length & 1{,}000 \\
  discounting & 0.99 \\
  num\_envs & 512 \\
  batch\_size & 512 \\
  policy\_lr & 3e-4 \\
  critic\_lr & 3e-4 \\
  contrastive\_loss\_function & symmetric InfoNCE \\
  energy\_function & L2 \\
  logsumexp\_penalty & 0.1 \\
  network depth & 16 \\
  network width & 256 \\
  representation dimension & 64 \\
  \rowcolor{blue!10} manifold & hyperboloid \\
  \rowcolor{blue!10} curvature $c$ & 0.5 \\
  \rowcolor{blue!10} hyperbolic layers & 1 (HTC~\citep{Yang2024Hypformer}) \\
  \rowcolor{blue!10} hyperbolic energy\_function & geodesic, $-d_c/\tau$ \\
  num\_epochs (one evaluation each) & \cellcolor{red!10} 200 (humanoid: 600) \\
  num\_eval\_envs & \cellcolor{red!10} 32 \\
  \bottomrule
  \end{tabular}
  \end{table}

\clearpage

\subsection{Offline CRL Hyperparameters}\label{app:arch_hparams:hparams_offline}

We adapt our online agent in two ways for the offline setting. First, we use the Minkowski inner product as an energy function instead of the geodesic distance. This mirrors offline Euclidean agents, which also perform better with a dot-product energy than with a distance energy~\citep{Park2025OGBench}. For a state-action embedding $\bm{x}$ and a goal embedding $\bm{y}$ on the hyperboloid $\hyperboloid$, the geodesic distance $d_{\hyperboloid}(\bm{x},\bm{y})$ strictly decreases as the Minkowski inner product $\minner{\bm{x}}{\bm{y}}$ increases (c.f.~Appendix~\ref{app:hyperbolic:hyp_ops}). Therefore, both energies rank goals in the same order, and the critic keeps its hyperbolic geometry. Second, we use a hybrid hyperbolic actor~\citep{Salemohamed2023HyperbolicContinuousControl} and regularize it with Hyper++~\citep{Klein2026HyperPP}. We hypothesize that the hyperbolic actor is stable for offline GCRL due to the absence of input nonstationarity.

  \begin{table}[h]
  \caption{\textbf{Offline (OGBench) hyperparameters.} Unshaded rows are the OGBench reference CRL configuration
  \citet{Park2025OGBench} and are shared by the Euclidean and the hyperbolic agent.
  \colorbox{blue!10}{Blue} rows exist only for the hyperbolic agent. \colorbox{red!10}{Red} rows differ from the reference.}
  \label{tab:hparams_offline}
  \centering
  \begin{tabular}{ll}
  \toprule
  Hyperparameter & Value \\
  \midrule
  train\_steps & 1M \\
  batch\_size & 1{,}024 \\
  optimizer / learning rate & Adam, 3e-4 \\
  actor MLP & (512, 512, 512), GELU, no LayerNorm \\
  critic MLP ($\phi$, $\psi$) & (512, 512, 512), GELU, LayerNorm \\
  representation dimension & 512 \\
  critic ensemble & 2 (min over members in the actor loss) \\
  discount & 0.99 (humanoidmaze, antmaze-giant: 0.995) \\
  contrastive loss & binary NCE \\
  energy\_function & $\phi^\top\psi / \sqrt{512}$ \\
  value goals (cur, traj, geom, rand) & (0, 0, 1, 0) \\
  actor goals (cur, traj, geom, rand) & navigate (0, 1, 0, 0); stitch (0, 0.5, 0, 0.5) \\
  actor loss & DDPG+BC, $\alpha = 0.1$, constant std \\
  \rowcolor{blue!10} manifold & hyperboloid \\
  \rowcolor{blue!10} critic curvature $c$ & 0.1 \\
  \rowcolor{blue!10} critic hyperbolic layers & 1 (HTC~\citep{Yang2024Hypformer}) per encoder \\
  \rowcolor{blue!10} hyperbolic energy\_function & Minkowski, $\langle\phi,\psi\rangle_{\mathcal{L}}/\tau$, $\tau = 1$ \\
  \rowcolor{blue!10} actor curvature $c_{\pi}$ & 0.5 \\
  \rowcolor{blue!10} actor hyperbolic layers & 1 (HTC), width 64 \\
  eval\_interval & 100k steps (11 evaluations incl.\ step 0; score = mean of the last 3) \\
  eval\_tasks & 5 \\
  eval\_episodes per task & \cellcolor{red!10} 20 \\
  \bottomrule
  \end{tabular}
  \end{table}

\clearpage

\section{Additional Results}\label{app:additional_results}

\subsection{Full Curvature Ablation}\label{app:additional_results:curvature_full}

\begin{figure}[h]
    \centering
        \includegraphics[width=0.8\linewidth]{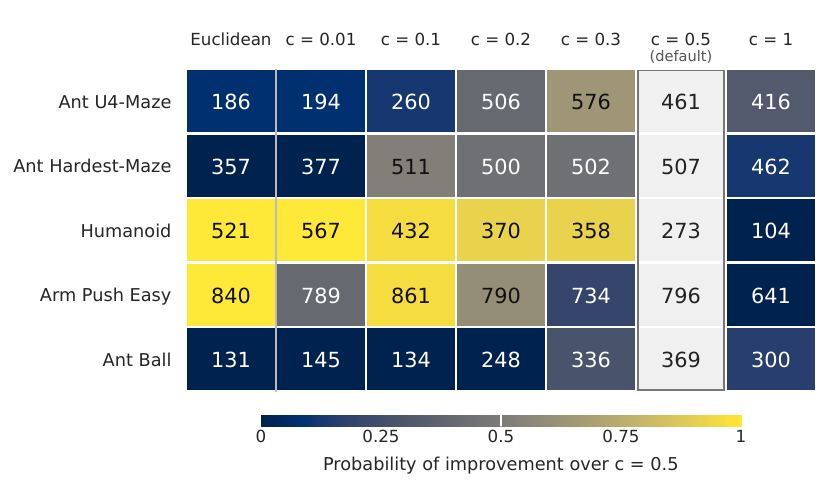}
        \caption{\textbf{Full curvature ablation.} Same Figure as Figure~\ref{fig:curvature_ablation}, including Ant Ball and the Euclidean agent as baseline. No single curvature setting $c$ works best across all environments, and the best $c$ depends on the environment. Hyperbolic agents with the curvature set per environment match Euclidean performance on Humanoid and Arm Push. Mean over the final 10 evaluation epochs over five seeds.} 
        \label{app:fig:curvature_ablation}
        \vspace{-3mm}
\end{figure}

Figure~\ref{app:fig:curvature_ablation} compares hyperbolic agents at several curvatures $c$ against the Euclidean agent across environments, including Ant Ball. On Humanoid and Arm Push, the Euclidean agent outperforms our default $c=0.5$ with high probability~\citep{Agarwal2021RliableStatisticalPrecipice}. However, on both, there exists a curvature that allows the hyperbolic agent to match the Euclidean steps-at-goal. No single curvature is best on all environments.

\clearpage

\subsection{Behavior on Ant Ball}\label{app:additional_results:ant_qualitative}

\begin{figure}[h]
    \centering
        \includegraphics[width=\linewidth]{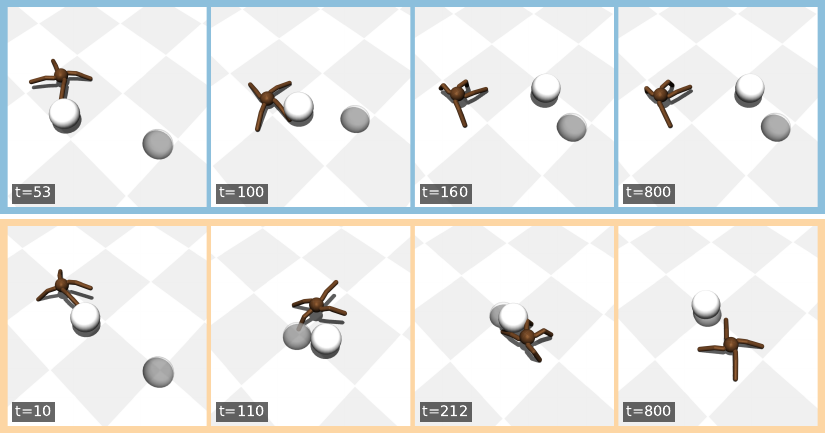}
        \caption{\textbf{Behavior on Ant Ball.}} 
        \label{app:fig:ant_ball_behavior}
        \vspace{-3mm}
\end{figure}

Figure~\ref{app:fig:ant_ball_behavior} shows typical failure and success behavior on Ant Ball. The Euclidean agent (top row) reaches the ball and pushes it once in roughly the target direction, but then the ball rolls off course and comes to rest 1.5 m from the target after step 118. The agent never returns to the ball after this push. For the rest of the episode, it moves around without heading toward the ball, the target, or a pushing position, and it ends 3 m from the ball. The hyperbolic agent (bottom row) instead pushes the ball 1.0 m past the target at step 110, circles to the far side, and pushes it back. From step 264 onward, the ball stays inside the goal region. We consistently observe this pattern across  640 evaluation episodes. 

\subsection{Full Offline Results Table}\label{app:additional_results:offline_table}

  \begin{table}[h]
  \centering
  \caption{With a Euclidean actor, the hyperbolic critic reaches a higher mean success rate than Euclidean CRL on seven of eight tasks, and adding the hyperbolic actor raises it further on seven of eight. Entries are mean OGBench success rates (\%) $\pm$ standard deviation over 8 seeds. \textbf{Bold} denotes hyperbolic entries whose 95\% bootstrap CI does not overlap that of Euclidean CRL. The last column is the full offline hyperbolic agent, whose hybrid hyperbolic actor is regularized with Hyper++~\citep{Klein2026HyperPP}.}
  \label{tab:offline-actor-ablation}
  \small
  \begin{tabular}{lccc}
  \toprule
   & & \multicolumn{2}{c}{Hyperbolic critic} \\
  \cmidrule(lr){3-4}
  Task & Euclidean CRL & Euclidean actor & Hyperbolic actor \\
  \midrule
  \multicolumn{4}{l}{\textit{AntMaze}} \\
  \quad Giant navigate  & 16.9 {\scriptsize$\pm$ 4.8} & \textbf{28.0} {\scriptsize$\pm$ 2.3} & \textbf{38.7} {\scriptsize$\pm$ 4.6} \\
  \quad Medium stitch   & 43.7 {\scriptsize$\pm$ 8.7} & \textbf{80.2} {\scriptsize$\pm$ 6.8} & \textbf{77.2} {\scriptsize$\pm$ 7.4} \\
  \quad Large stitch    & 17.6 {\scriptsize$\pm$ 3.2} & 21.2 {\scriptsize$\pm$ 6.6} & 22.6 {\scriptsize$\pm$ 7.5} \\
  \midrule
  \multicolumn{4}{l}{\textit{HumanoidMaze}} \\
  \quad Medium navigate & 60.2 {\scriptsize$\pm$ 3.0} & \textbf{66.1} {\scriptsize$\pm$ 4.2} & \textbf{68.2} {\scriptsize$\pm$ 3.9} \\
  \quad Large navigate  & 23.5 {\scriptsize$\pm$ 6.0} & 20.5 {\scriptsize$\pm$ 5.4} & 24.9 {\scriptsize$\pm$ 6.9} \\
  \quad Giant navigate  &  5.9 {\scriptsize$\pm$ 2.1} & \textbf{10.7} {\scriptsize$\pm$ 2.6} & \textbf{16.8} {\scriptsize$\pm$ 2.9} \\
  \quad Medium stitch   & 56.0 {\scriptsize$\pm$ 3.2} & \textbf{63.9} {\scriptsize$\pm$ 3.6} & \textbf{68.0} {\scriptsize$\pm$ 2.2} \\
  \quad Large stitch    &  8.9 {\scriptsize$\pm$ 4.2} & \textbf{14.4} {\scriptsize$\pm$ 1.4} & \textbf{17.7} {\scriptsize$\pm$ 2.0} \\
  \bottomrule
  \end{tabular}
  \end{table}

\clearpage

\section{Wall-clock Timings}\label{app:wall_clock}

\begin{figure}[t]
  \centering
  \begin{subfigure}[t]{0.32\linewidth}
    \includegraphics[width=\linewidth]{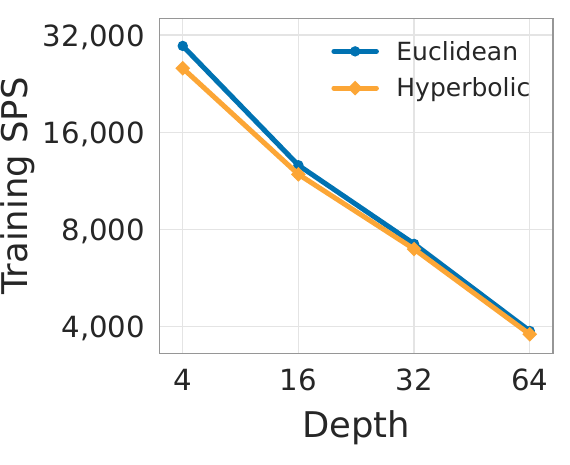}
    \caption{Training throughput.}
    \label{fig:u4-runtime-sps}
  \end{subfigure}\hfill
  \begin{subfigure}[t]{0.32\linewidth}
    \includegraphics[width=\linewidth]{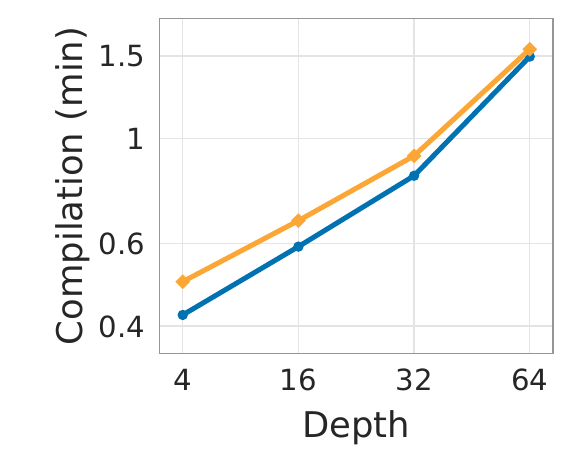}
    \caption{Compilation overhead.}
    \label{fig:u4-runtime-compile}
  \end{subfigure}\hfill
  \begin{subfigure}[t]{0.32\linewidth}
    \includegraphics[width=\linewidth]{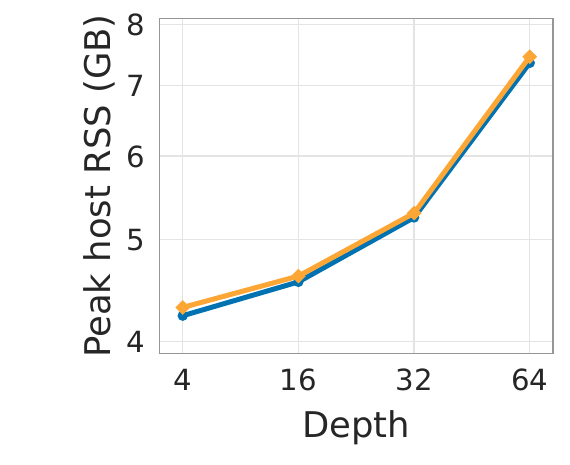}
    \caption{Peak host memory.}
    \label{fig:u4-runtime-memory}
  \end{subfigure}
  \caption{\textbf{Runtime on Ant U4-Maze versus network depth on one H100 GPU.} 
  \textbf{\ref{fig:u4-runtime-sps})} Median SPS over retained epochs. 
   \textbf{\ref{fig:u4-runtime-compile})} JAX compilation time. 
   \textbf{\ref{fig:u4-runtime-memory})} Peak host memory. All vertical axes are logarithmic.}
  \label{fig:u4-runtime}
\end{figure}

In this section, we measure training throughput, compilation overhead, and peak host memory on Ant U4-Maze, Humanoid U-Maze, and Humanoid Big-Maze. All runs use the hyperparameters specified in Appendix~\ref{app:arch_hparams:hparams_online}. Throughput is measured in environment steps per second (SPS) within a training epoch. It covers environment interaction, replay sampling, parameter updates, and training diagnostics. It excludes setup, replay prefill, and evaluation. We discard the first epoch and report the median SPS over the remaining epochs. For depths up to 64, we measure nine epochs of five training steps. For deeper networks, we use four epochs of four training steps. We estimate compilation overhead as the first epoch's duration minus the median duration of later epochs, including all first-call costs, such as tracing and autotuning, not only compiler time. Peak host memory is the maximum resident set size (RSS) of the process, reported by \texttt{/usr/bin/time -v} in decimal GB.

Figure~\ref{fig:u4-runtime} reports these metrics for Ant U4-Maze, and Table~\ref{tab:humanoid-runtime} reports them for Humanoid U-Maze and Humanoid Big-Maze. Figures~\ref{fig:u4-runtime-sps} and~\ref{fig:u4-runtime-compile} show that the hyperbolic critic head lowers throughput and modestly increases compilation overhead. For SPS, the maximum overhead is 14.71\% at depth 4, but this quickly drops for deeper networks. At our main evaluation setting using depth 16, hyperbolic agents achieve 6.27\% lower throughput. Peak host memory is nearly unchanged, see Figure~\ref{fig:u4-runtime-memory}. In contrast, deeper Euclidean networks are far more expensive: going from depth 64 to 512 leads to several times lower throughput and raises compilation overhead and peak host memory by more than an order of magnitude, as shown in Table~\ref{tab:humanoid-runtime}.

For a budget of $N$ environment steps, the estimated time in hours is
\begin{equation*}
  \hat{T}(N) = \frac{N \, t_{\text{step}}}{31{,}744 \times 3{,}600}~,  
\end{equation*}
where $t_{\text{step}}$ is the median epoch duration in seconds divided by the number of training steps per epoch, and $31{,}744$ the number of transitions collected per iteration. Thus, $\hat{T}(N)$ is the number of training steps needed to collect $N$ transitions times the measured seconds per step, converted to hours. It excludes compilation, setup, prefill, evaluation, and checkpointing. 

  \begin{table}[t]
  \centering
  \caption{\textbf{Runtime on the Humanoid mazes on one H100 GPU.} SPS are measured as median environment steps per second, excluding evaluation. We estimate the compile time as first-epoch duration minus median later-epoch duration. Peak RSS does not refer to GPU VRAM. The hyperbolic head adds 8{,}448 parameters irrespective of depth.}
  \label{tab:humanoid-runtime}
  \setlength{\tabcolsep}{4pt}
  \begin{tabular}{lrr S[table-format=3.1] S[table-format=4.0,group-separator={,},group-minimum-digits=4]
    S[table-format=2.2] S[table-format=3.1]}
  \toprule
  Method & Actor & Critic & {Params} & {Median SPS} & {Est. compile} & {Peak RSS} \\
   & depth & depth & {(M)} & {($\uparrow$)} & {(min, $\downarrow$)} & {(GB, $\downarrow$)} \\
  \midrule
  \multicolumn{7}{l}{\textit{Humanoid U-Maze}} \\
  Hyperbolic & 64 & 64 & 12.9 & 3722 & 1.62 & 7.6 \\
  Euclidean & 64 & 64 & 12.9 & 3788 & 1.54 & 7.6 \\
  Euclidean & 512 & 512 & 102.0 & 512 & 32.52 & 83.2 \\
  Euclidean & 512 & 1024 & 169.9 & 303 & 92.79 & 299.4 \\
  \midrule
  \multicolumn{7}{l}{\textit{Humanoid Big-Maze}} \\
  Hyperbolic & 64 & 64 & 12.9 & 3635 & 1.64 & 7.6 \\
  Euclidean & 64 & 64 & 12.9 & 3686 & 1.54 & 7.6 \\
  Euclidean & 512 & 512 & 102.0 & 510 & 32.44 & 83.1 \\
  \bottomrule
  \end{tabular}
  \end{table}

%% file: tikz/gromov_product.tikz
\begin{tikzpicture}[
    edge/.style={line width=1.7pt, draw=black!65},
    vertex/.style={circle, fill=blue!70, draw=black,
                   line width=0.8pt, minimum size=7pt, inner sep=0pt},
    branch/.style={circle, fill=black, minimum size=5pt, inner sep=0pt}
]

\coordinate (w1) at (0,0);
\coordinate (x1) at (3.5,2.2);
\coordinate (z1) at (3.5,-1.8);

\draw[edge] (w1) -- (x1)
    node[midway, above left] {$d(w,x)$};
\draw[edge] (w1) -- (z1)
    node[midway, below left] {$d(w,z)$};
\draw[edge] (x1) -- (z1)
    node[midway, right] {$d(x,z)$};

\node[vertex, label=left:{$w$}] at (w1) {};
\node[vertex, label=above:{$x$}] at (x1) {};
\node[vertex, label=below:{$z$}] at (z1) {};

\coordinate (p)  at (2+8,0);
\coordinate (w2) at (2+6.1,0);
\coordinate (x2) at (2+10,1.8);
\coordinate (z2) at (2+10,-1.8);

\draw[edge] (w2) -- (p)
    node[midway, below=5pt] {$(x,z)_w$};

\draw[edge] (p) -- (x2)
    node[midway, above left=2pt] {$(w,z)_x$};

\draw[edge] (p) -- (z2)
    node[midway, below left=2pt] {$(w,x)_z$};

\node[vertex, label=left:{$w$}] at (w2) {};
\node[vertex, label=above:{$x$}] at (x2) {};
\node[vertex, label=below:{$z$}] at (z2) {};
\node[branch, label=above left:{$p$}] at (p) {};

\end{tikzpicture}

%% file: tikz/model_architecture.tikz
\begingroup%
\definecolor{oxblue}{HTML}{002147}%
\definecolor{Tgrey}{HTML}{7a7977}%
\definecolor{blue_acc}{HTML}{4095c4}%
\definecolor{smaragd_acc}{HTML}{41b696}%
\definecolor{mlrblue}{HTML}{96c5df}%
\definecolor{geored}{HTML}{C54746}%
\def\archBl{3.40cm}%
\def\archBh{0.682cm}%
\begin{tikzpicture}[
  x=1cm, y=1cm, font=\small, >=stealth,
  pblock/.style={draw=black, line width=1.2pt, rounded corners=3pt, fill=Tgrey!40!white,
                 text=black, font=\small\bfseries, minimum width=\archBl,
                 minimum height=\archBh, align=center, rotate=90},
  hblock/.style={draw=black, line width=1.2pt, rounded corners=3pt, fill=mlrblue,
                 text=black, font=\small\bfseries, minimum width=\archBl,
                 minimum height=\archBh, align=center, rotate=90},
  flow/.style={->, thick, color=oxblue!70},
]

\draw[oxblue!40, rounded corners=2pt] (-2.90,-1.667) rectangle (2.90,1.667);
\draw[oxblue!30, dashed, fill=Tgrey!15] (-2.80,-1.567) rectangle (2.80,1.567);
\node[text=oxblue!60, font=\small\bfseries, align=center] at (0,0)
    {\includegraphics[width=5.60cm,height=3.134cm]{figures/online_envs.png}};
\coordinate (imgE) at (2.90,0);

\filldraw[fill=Tgrey!40!white, draw=black, line width=1.2pt]
  (3.75,1.60) -- (6.15,0.90) -- (6.15,-0.90) -- (3.75,-1.60) -- cycle;
\node[align=center, text=black, font=\small\bfseries] at (4.95,0)
  {Euclidean\\ResNet};

\node[pblock] (exp) at (7.55,0) {Exponential Map};

\node[hblock] (hlin) at (9.55,0) {Hyperboloid Linear};

\draw[black, line width=1.2pt, rounded corners=5pt]
  (10.80,-1.93) rectangle (15.50,1.93);
\begin{scope}[shift={(13.15,-3.0996)}, x=0.656cm, y=0.8856cm,
              line cap=round, line join=round]
  \fill[mlrblue]
    (3.2,4.6) .. controls (3.2,4.15) and (1.6,1.55) .. (0,1.55)
              .. controls (-1.6,1.55) and (-3.2,4.15) .. (-3.2,4.6)
              arc[start angle=180, end angle=360, x radius=3.2, y radius=0.85]
    -- cycle;
  \draw[oxblue, line width=0.55pt]
    (3.2,4.6) .. controls (3.2,4.15) and (1.6,1.55) .. (0,1.55)
              .. controls (-1.6,1.55) and (-3.2,4.15) .. (-3.2,4.6);
  \draw[mlrblue!80, line width=0.55pt]
    (3.2,4.6) arc[start angle=0, end angle=180, x radius=3.2, y radius=0.85];
  \fill[mlrblue!80] (0,4.6) ellipse[x radius=3.2, y radius=0.85];
  \draw[oxblue!60, line width=1.0pt]
    (-3.2,4.6) arc[start angle=180, end angle=360, x radius=3.2, y radius=0.85];
  \draw[geored, line width=1.4pt]
    (-1.65,3.05) .. controls (-0.95,2.35) and (0.85,2.50) .. (1.90,3.45);
  \fill[geored] (-1.65,3.05) circle[radius=1.7pt];
  \fill[geored] ( 1.90,3.45) circle[radius=1.7pt];
  \node[text=geored, font=\small\bfseries, anchor=south east, inner sep=1.75pt]
    at (-1.66,3.09) {$\bm{x}$};
  \node[text=geored, font=\small\bfseries, anchor=south west, inner sep=1.75pt]
    at ( 1.91,3.49) {$\bm{y}$};
\end{scope}

\draw[flow] (imgE)       -- (3.75,0);
\draw[flow] (6.15,0)     -- (exp.north);   %
\draw[flow] (exp.south)  -- (hlin.north);
\draw[flow] (hlin.south) -- (10.80,0);

\begin{scope}[on background layer]
  \draw[smaragd_acc!80, dashed, very thick, rounded corners=5pt,
        fill=smaragd_acc, fill opacity=0.15] (3.35,2.25) rectangle (6.55,-2.25);
  \draw[blue_acc!80, dashed, very thick, rounded corners=5pt,
        fill=blue_acc, fill opacity=0.12] (8.65,2.25) rectangle (15.90,-2.25);
\end{scope}
\node[text=smaragd_acc!80, font=\small\bfseries, anchor=south west] at (3.35,2.25) {Euclidean};
\node[text=blue_acc!80,    font=\small\bfseries, anchor=south west] at (8.65,2.25) {Hyperbolic};

\end{tikzpicture}%
\endgroup%